\pdfoutput=1

\documentclass[11pt]{article}

\usepackage[]{acl}

\usepackage{times}
\usepackage{latexsym}

\usepackage[T1]{fontenc}

\usepackage[utf8]{inputenc}
\usepackage{enumitem}

\usepackage{microtype}
\usepackage{longtable}
\usepackage{booktabs}

\usepackage[]{xcolor} 
\usepackage{graphicx}
\usepackage{multirow}
\usepackage[breakable,skins]{tcolorbox}

\usepackage{inconsolata}
\usepackage{amsmath,amssymb}

\newtcolorbox{promptbox}[1][]{
    breakable,
    colback=gray!10,    
    colframe=gray!20,     
    title=#1,           
    fonttitle=\bfseries,
    boxrule=0.5mm,      
    arc=2mm,            
    outer arc=2mm,      
    coltitle=black,     
    enhanced,
}
\usepackage{lscape}
\usepackage{scrextend}
\title{A Modular Approach for Clinical SLMs Driven by Synthetic Data with Pre-Instruction Tuning, Model Merging, and Clinical-Tasks Alignment}

\author{
  \vspace{-0.5cm}\\{\bf Jean-Philippe Corbeil}\textsuperscript{1}\thanks{Corresponding author: \textit{jcorbeil@microsoft.com}},
  {\bf Amin Dada}\textsuperscript{4},
  {\bf Jean-Michel Attendu}\textsuperscript{1},
  {\bf Asma Ben Abacha}\textsuperscript{1}\\
  {\bf Alessandro Sordoni}\textsuperscript{2,3}, 
  {\bf Lucas Caccia}\textsuperscript{2},
  {\bf François Beaulieu}\textsuperscript{1},
  {\bf Thomas Lin}\textsuperscript{1}\\
  {\bf Jens Kleesiek}\textsuperscript{4}\thanks{Other affiliations:
 Cancer Research Center Cologne Essen (CCCE), German Cancer Consortium (DKTK, Partner site Essen) and Department of Physics of TU Dortmund (Dortmund, Germany).},
  {\bf Paul Vozila}\textsuperscript{1}\\\\
  \textsuperscript{1}Microsoft Healthcare \& Life Sciences\ \ \ 
  \textsuperscript{2}Microsoft Research Montréal, Canada \\
  \textsuperscript{3}Mila, Université de Montréal, Canada \ \ 
  \textsuperscript{4}IKIM, University Hospital Essen, Germany
}

\begin{document}
\maketitle

\begin{abstract}
High computation costs and latency of large language models such as GPT-4 have limited their deployment in clinical settings. Small language models (SLMs) offer a cost-effective alternative, but their limited capacity requires biomedical domain adaptation, which remains challenging. An additional bottleneck is the unavailability and high sensitivity of clinical data. To address these challenges, we propose a novel framework for adapting SLMs into high-performing clinical models. We introduce the \textbf{MediPhi} collection of 3.8B-parameter SLMs developed with our novel framework: pre-instruction tuning of experts on relevant medical and clinical corpora (PMC, Medical Guideline, MedWiki, etc.), model merging, and clinical-tasks alignment. To cover most clinical tasks, we extended the CLUE benchmark to CLUE+, doubling its size. Our expert models deliver relative improvements on this benchmark over the base model without any task-specific fine-tuning: 64.3\% on medical entities, 49.5\% on radiology reports, and 44\% on ICD-10 coding (outperforming GPT-4-0125 by 14\%). We unify the expert models into MediPhi via model merging, preserving gains across benchmarks. Furthermore, we built the \textbf{MediFlow} collection, a synthetic dataset of 2.5 million high-quality instructions on 14 medical NLP tasks, 98 fine-grained document types, and JSON format support. Alignment of MediPhi using supervised fine-tuning and direct preference optimization achieves further gains of 18.9\% on average.
\end{abstract}

\begin{figure}[!h]
\centering
\includegraphics[width=\linewidth]{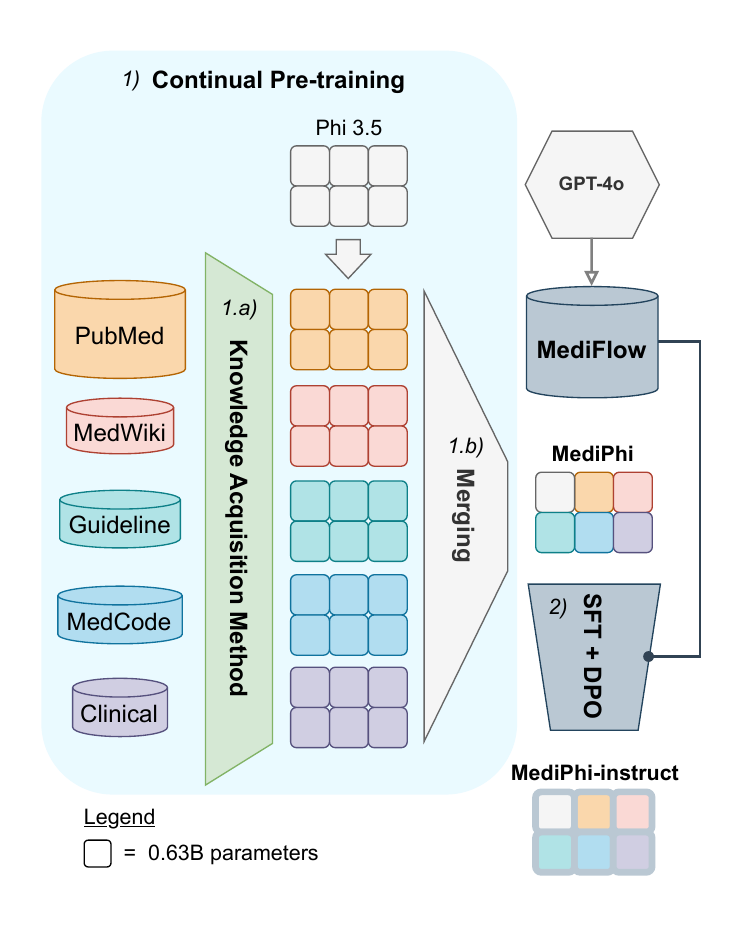}
\caption{Our approach in two steps: 1) continual pre-training; 2) alignment. 1) Starting from~\textit{Phi3.5 mini}: 1.a) We leverage \emph{knowledge acquisition methods} such as pre-instruction tuning on diverse medical and clinical corpora to obtain domain-specific experts. 1.b) We recur to \emph{model merging} to merge experts first with the base model to combine new knowledge as well as recover skills degraded by the previous step, then together to form an unified model, MediPhi. 2) We generate MediFlow, a synthetic instruction dataset MediFlow for clinical tasks. We align our model on MediFlow using Supervised Fine-Tuning (SFT) and Direct Preference Optimization (DPO) and obtain MediPhi-Instruct. \textit{Segmentation of model parameters as equal block size are only for illustrative purposes}.}
\label{fig:global_diagram}
\vspace{-0.5cm}
\end{figure}

\section{Introduction}
Advances in natural language processing (NLP) have enabled large language models (LLMs) like GPT-4 to excel in medical tasks, especially medical exams \cite{nori2023can,abacha2024medec,nori2024medprompt}. However, their deployment in clinical settings\footnote{We define the medical field as encompassing medical knowledge (e.g. anatomy, genetics, biology), while the clinical field is related to direct patient care and healthcare practice (e.g. doctor-patient dialog, discharge summary, and clinical note).} faces many challenges, including high latency and cost \cite{yang2023large,dennstadt2025implementing}. As LLM progress faces diminishing returns from scaling \cite{udandarao2024plateauing,longpre2024consent,muennighoff2023scaling,Villalobos2022runoutdata}, specialized small language models (SLMs) provide a viable alternative \cite{sardana2023beyondchinchilla}, when optimized for domain-specific performance, lower computational requirements, and real-world clinical integration.

Developing high-quality clinical language models is hindered by the unavailability of clinical data which are sensitive and tightly licensed, e.g. protected health information under HIPAA. Current medical LLMs perform well on multiple-choice question datasets but struggle with real-world clinical complexities \cite{dada2024clue,chen2024clinicalbench,Liu2024benchmark,jeong2024medical,jeong2024limited}. Furthermore, both the inaccessibility of clinical data and the misalignment of current continual pre-training methods for clinical tasks are critical limitations in the context of SLMs, which have limited capacity. Addressing these gaps require innovative training strategies tailored to small models.

This work introduces a modular framework for building high-performance medical SLMs, leveraging pre-instruction tuning (PIT) \cite{jiang2024pit}, model merging, and clinical alignment. Using pre-instruction tuning, we adapt \textit{Phi3.5 mini} with 3.8B parameters \cite{abdin2024phi3} into experts trained on diverse medical and clinical corpora. We unify these models through model merging into one SLM which preserves benchmark improvements. We complete training by aligning the model with MediFlow, a new synthetic instruction dataset on clinical tasks. A representation of this approach is given in Figure \ref{fig:global_diagram}.

\noindent Our contributions include:
\begin{itemize}
    \item Introducing the \textbf{MediPhi family}, the first collection of high-performance SLMs for medical and clinical applications under a commercially permissive license\footnote{\label{fn:release}TBD}. The collection includes both a generalist expert model and specialized variants. Additionally, we release synthetic validation sets designed to guide the model merging algorithm, facilitating reproducibility, and enabling future integration of other clinical expert models.
    \item Releasing of \textbf{MediFlow collection} of 2.5 million high-quality synthetic instructions generated with a GPT-4o based agentic pipeline, also under a commercially permissive license\footref{fn:release}. This dataset for the clinical domain contains 14 tasks categories, 98 fine-grained input documents, 6 difficulty levels, and 2 output formats (JSON or plain text), filling a gap in the clinical NLP resources. 
    \item An extension of CLUE to the \textbf{CLUE+ benchmark} by doubling its size from 6 to 12 datasets including complementary clinical tasks and input documents, allowing a comprehensive evaluation of medical and clinical capabilities of language models (e.g. radiology reports, medications, medical error detection, doctor-patient dialog summarization, information extraction of social determinants of health, and medical coding).
    \item A demonstration of the \textbf{effectiveness of pre-instruction tuning} for medical domain adaptation extending the method beyond question-answering with named entity recognition, relation extraction, and summarization.
    \item A \textbf{study on ICD10CM medical coding}, in terms of domain adaptation and benchmarking of medical models, with relative improvements up to 44\% over the base model, surpassing GPT-4-0125 by 14\%.
\end{itemize}

\section{Previous Work}
\subsection{Medical Large Language Models}
Researchers have developed various open-weight LLMs with diverse capabilities and research licenses in the medical NLP domain. Examples include ClinicalCamel \cite{toma2023clinical}, Med42 \cite{med42}, PMC-Llama \cite{wu2023pmcllama}, BioMedGPT \cite{zhang2023biomedgpt}, Meditron \cite{chen2023meditron}, BioMistral \cite{labrak2024biomistral} and Asclepius \cite{kweon2024asclepius}. Most recent medical LLMs \cite{chen2023meditron,christophe2024med42v2,gururajan2024aloe,OpenBioLLMs} are based on Llama3 \cite{dubey2024llama3} --- 8B and 70B parameters. Google also trained their own  medical LLM named Med-PaLM 2 \cite{singhal2023large} with 340B parameters, which is not publicly available.

\subsection{Medical Instruction Tuning}
Recently, authors have released instruction-tuned models for the medical domain: Aloe \cite{gururajan2024aloe}, Hippocrates \cite{acikgoz2024hippocrates}, and Med42 v2 \cite{christophe2024med42v2,christophe2024beyond}. The alignment phase of these three models based on only supervised fine-tuning includes similar instruction datasets such as medical question-answering databases, non-medical alignment data (e.g. UltraChat by \citeauthor{ding2023ultrachat}), and benchmark training sets --- e.g. MedQA \cite{jin2021medqa} and PubMedQA \cite{jin2019pubmedqa}. While this mix of data has contributed to improvements, it also introduces imbalances in task and document coverage, as well as in-distribution evaluations in some cases, i.e. closer to evaluation in fine-tuning setting instead of zero-shot/few-shot setting.

\subsection{Synthetic Instructions}
Phi 1 \& 2 \cite{gunasekar2023textbooks,li2023textbooks2}, with 1.3B and 2.7B parameters, respectively, showed strong reasoning performance using synthetic textbook-like data. Phi-3 and Phi-3.5 mini \cite{abdin2024phi3} scaled model size to 3.8B parameters and topic coverage. Phi4 \cite{abdin2024phi4} at 14B parameters is trained using an iterative data generation process. Similarly, Orca \cite{mukherjee2023orca,mitra2024agentinstruct} focused on reasoning data, while UltraChat \cite{ding2023ultrachat} targeted multi-turn prompts. \citet{zhang2023alpacare} generated 52,000 medical question-answering instructions based on forms filled by experts. In the clinical domain, \citet{kweon2024asclepius} generated 158k short question-answering instructions for 8 tasks with synthetic clinical documents as inputs, seeded from the PMC-Patients corpus \cite{zhao2022pmcpatients}. MagPie \cite{xu2024magpie} proposed the generation of synthetic instructions in general domains using LLMs' chat template.

\subsection{Model Merging}
Model Soup \cite{wortsman2022modelsoup} demonstrated improvements over single checkpoint evaluation by averaging training checkpoints, leading to methods like spherical linear interpolation (SLERP), enabled by linear-mode connectivity \cite{frankle2020lmc1,mirzadeh2020lmc2}. To extend to multiple models, authors have proposed \textit{task arithmetic} \cite{ilharco2022taskvec} while techniques such as TIES \cite{yadav2024ties}, DARE \cite{yu2024mario}, and BreadCrumbs \cite{davari2024breadcrumbs} address the parameter interference issue. \citet{hammoud2024modelmergesafety} highlighted the importance of validation sets for an optimal merge. Large-scale experiments \cite{yadav2024mergescale} showed the effectiveness of merging multiple experts, especially from large instruction models. Merging has also proven as effective as data-mix strategies during pre-training \cite{ahmadian2024mixdata,na2024scalable}.

\subsection{Clinical Benchmarking}
Medical LLMs are commonly benchmarked on medical knowledge through multiple-choice datasets such as MedQA \cite{jin2021medqa}, MedMCQA \cite{pal2022medmcqa}, PubMedQA \cite{jin2019pubmedqa}, and MMLU-medical \cite{hendrycks2020mmlu}. However, recent studies \cite{Liu2024benchmark,dada2024clue,chen2024clinicalbench,jeong2024medical,jeong2024limited} revealed a gap on the clinical domain from medical LLMs. 

\section{Methodology}
Our method for clinical SLMs as illustrated in Figure \ref{fig:global_diagram} includes two steps: 1) continual pre-training composed of 1.a) domain knowledge acquisition methods and 1.b) model merging, and 2) post-training with supervised fine-tuning and direct preference optimization on generated synthetic data.

\subsection{Continual Pre-Training}

\subsubsection{Datasets}
In Table \ref{tab:dataset_info}, we list the medical and clinical corpora with permissive licenses, used to adapt our SLM into five experts. We separate these into five groups: \textit{PubMed}, \textit{Clinical}, \textit{MedCode}, \textit{Guidelines}, and \textit{MedWiki}. We describe these dataset groups with their licensing status in detail in Appendix \ref{sec:dapt_datasets}.

\begin{table}[h!]
\caption{Medical and clinical data sources separated into five groups: \textit{PubMed}, \textit{Clinical}, \textit{MedCode}, \textit{Guidelines}, and \textit{MedWiki}.}
\centering
\setlength{\tabcolsep}{0.4em}
\renewcommand{\arraystretch}{1.3}
\small
\begin{tabular}{lllcc}
\toprule
\textbf{\scriptsize Groups} & \textbf{\scriptsize Source} & \textbf{\scriptsize Document Type} & \textbf{\scriptsize \#Docs} & \textbf{\scriptsize \#Tokens} \\
\midrule
\multirow{2}{*}{\tiny \textbf{PubMed}} & {\scriptsize PMC} & {\scriptsize Scientific Articles} & {\scriptsize 3.8M} & {\scriptsize 42B} \\
    & {\scriptsize PMC Abstract} & {\scriptsize Scientific Abstracts} & {\scriptsize 36M} & {\scriptsize 6B} \\
\hline
\multirow{4}{*}{\tiny \textbf{Clinical}} & {\scriptsize PMC-Patients} & {\scriptsize Patient summaries} & {\scriptsize 156k} & {\scriptsize 130M} \\
& {\scriptsize Asclepius} & {\scriptsize Clinical Documents} & {\scriptsize 80k} & {\scriptsize 44M} \\
& {\scriptsize NoteChat} & {\scriptsize Conversations} & {\scriptsize 80k} & {\scriptsize 72M} \\
 & {\scriptsize MTSamples} & {\scriptsize Clinical Documents}  & {\scriptsize 5k} & {\scriptsize 4M} \\
\hline
{\tiny \textbf{MedCode}} & {\scriptsize ICD9/10\&ATC} & {\scriptsize Webpages} &{\scriptsize 206k} & {\scriptsize 257M} \\
\hline
{\tiny \textbf{Guidelines}} & {\scriptsize Guidelines} & {\scriptsize Websites} & {\scriptsize 37k} & {\scriptsize 90M} \\
\hline
{\tiny \textbf{MedWiki}} & {\scriptsize MedWiki} & {\scriptsize Encyclopedia}  & {\scriptsize 80k} & {\scriptsize 80M} \\
\hline
\end{tabular}
\label{tab:dataset_info}
\end{table}

\subsubsection{Domain-Specific Pre-Training}
\label{sec:desc-cp}
We consider three methods to enhance domain knowledge of language models: domain adaptation pre-training (DAPT), textbook-like synthetic material (Explainer), and pre-instruction tuning (PIT). The latter showed considerable improvements in our experiments. We apply these techniques on the base model to obtain five experts in the medical and clinical domain from our five dataset groups.

\paragraph{DAPT}
\cite{ganin2015unsuperviseddomainadapt, gururangan2020dontspotpretraining} Domain Adaptation Pre-Training is a technique for adapting a deep learning model to a specific domain by performing next-token prediction on domain-specific corpora. The expert that is trained on the \textit{PubMed} group follows standard DAPT, because it is orders of magnitude larger than the others. We also train on all the data a \textit{DataMix} baseline using DAPT, which is outperformed by our approach. For this method, we trained models for one epoch with a cosine scheduler at a maximum learning rate of 5e-5.

\paragraph{Explainer}
\cite{gunasekar2023textbooks} Textbook-like material demonstrated the effectiveness of training SLMs on textbook-quality data generated by a strong LLM. Therefore, we generate a textbook-like "explainer" using GPT-4o for the \textit{MedCode} dataset group, for which the dense format of the webpages impedes model learning. For this method, we trained models for two epochs with a cosine scheduler at a maximum learning rate of 1e-4.

\paragraph{PIT}
\cite{jiang2024pit} Pre-Instruction Tuning\footnote{Despite its name, PIT does not involve explicit instruction data or chat-style formatting.} (PIT) demonstrated significant improvements over the conventional training paradigm, by first fine-tuning on instruction-like data, followed by training on a concatenation of the instruction data and pre-training corpus. 


PIT requires the generation of task data for each document in the corpus. While \citet{jiang2024pit} originally used question-answering (QA) as the primary task, we expand the approach to include summarization, named entity recognition, and relation extraction. We use GPT-4o to generate outputs for all four tasks on the ICD10CM subset of the MedCode dataset as an initial case study. Based on the results, we extend this process to the remaining four dataset groups --- Clinical, MedCode, Guidelines, and MedWiki --- to train multiple expert models.

Each expert undergoes sequential training in two phases. The first phase involves fine-tuning on the generated outputs from a single task for one epoch using a cosine scheduler (peak learning rate at 1e-4). If the task contains multiple elements, such as several question-answer pairs, they are concatenated into a single sequence, separated by end-of-sentence (EOS) tokens. In the second phase, the model is fine-tuned for two epochs with another cosine scheduler (peak learning rate at 3e-4) on the concatenation of the task data and the original documents, with EOS tokens acting as separators.

\begin{figure}[!ht]
\centering
\includegraphics[width=0.9\linewidth]{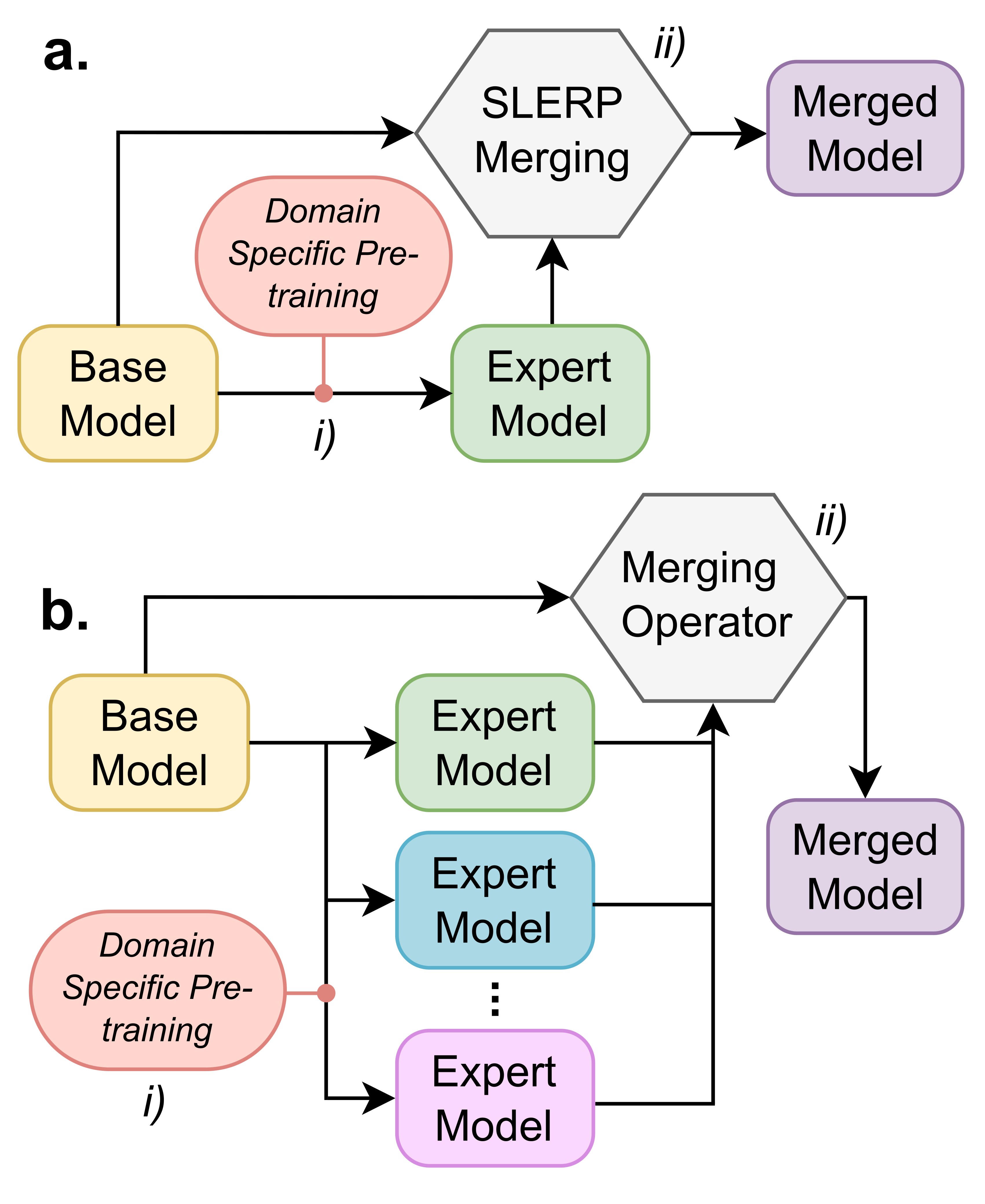}
\caption{Overview of model merging techniques. \textbf{a.} \textit{Domain-specific merging via SLERP}: An expert model is obtained by fine-tuning the base model on domain-specific data (step i). The expert and base models are then merged using spherical linear interpolation (SLERP) to produce the final merged model (step ii). \textbf{b.} \textit{Multi-expert merging}: Multiple expert models are independently derived from the base model via domain-specific pretraining (step i). These are then combined using a merging operator (e.g., Task Arithmetic, Ties, or BreadCrumbs) to produce a unified merged model (step ii).}
\label{fig:merging_diag}
\end{figure}

\subsubsection{Domain-Specific Model Merging}
We train five expert models specializing in different aspects of medical and clinical knowledge, each derived from a distinct dataset group: \textit{PubMed}, \textit{Clinical}, \textit{MedCode}, \textit{Guidelines}, and \textit{MedWiki}.

Following the success of BioMistral \cite{labrak2024biomistral}, the first approach involves merging each expert model individually with the base model using SLERP\footnote{A method optimized for merging two models, often yielding high-performing hybrids \cite{hammoud2024modelmergesafety,ahmadian2024mixdata,labrak2024biomistral}.} as in step \textit{a} of Figure \ref{fig:merging_diag}. We determined the merging proportions (10\%, 25\% or 50\% ) via validation sets (see below). We apply merging after PIT since these techniques demonstrate a synergistic effect. While PIT enhances domain-specific learning, it also leads to catastrophic forgetting --- degrading the model’s initial abilities such as instruction following, long context handling, and multilingual support \cite{scialom2022t0}. This issue is particularly evident in zero-shot and few-shot settings, where instruction-tuned models generally outperform their base counterparts \cite{longpre2023flan,zhang2023instruction}. Merging with the original instruction model after PIT mitigates these degradations, preserving general capabilities while maximizing domain adaptation\footnote{Alternative approaches, such as KL-divergence regularization in RLHF \cite{ouyang2022rlhf}, exist for maintaining generalization, but recent studies suggest model merging can further optimize reward alignment in RL settings \cite{rame2024warp}.}.

\subsubsection{Multi-Expert Merging into MediPhi}
Our second set of experiments combines all five experts into a unified SLM forming $MediPhi$ as in step \textit{b} of Figure \ref{fig:merging_diag}. Multi-model merging involves three primary techniques: Task Arithmetic, TIES, and BreadCrumbs. Given the vast configuration space of multi-model merges, we employ an evolutionary algorithm via MergeKit \cite{goddard2024mergekit} to optimize the merging process.

However, optimization on our benchmark is not feasible due to a lack of validation data and framework incompatibilities. To address this, we generate synthetic validation sets aligned with our benchmark tasks. Specifically, we prompt GPT-4o to create multiple-choice question sets covering 12 medical and clinical topics relevant to our benchmark (e.g., doctor-patient interactions, medical coding, discharge summaries). These sets maintain contextual consistency with our evaluation tasks. The evolutionary algorithm is set to terminate after 500 evaluations, guided by average accuracy on these validation sets. Further details about validation set creation procedures are provided in Appendix \ref{sec:valsets}.

\subsubsection{Evaluation Metrics}
To identify the top-performing model, we primarily measure the \textit{average accuracy} on CLUE+. However, we also consider important that the expert achieves uniform improvements, especially among experts with similar accuracies. For this purpose, we use $\#DG$, the number of datasets on which the model achieves gains, and CV $\Delta$, the \textit{coefficient of variations of gains/losses} as defined in equation \ref{eq:cvd}.

\begin{equation}
\label{eq:cvd}
    CV\ \Delta = \frac{\sqrt{\mathbb{E}_{d \sim \mathcal{D}}\left[\left(\delta_d - \mu_d\right)^2\right]}}{|\mu_d|}
\end{equation}

where $\delta_d$ is the expert accuracy minus the baseline one for the $d^{th}$ dataset in the benchmark $D$ and $\mu_d=\mathbb{E}_{d \sim \mathcal{D}}[\delta_d]$. A small CV $\Delta$ indicate uniform gains or losses across datasets, while a high value indicates gains or losses on a narrow subset of datasets.

\subsection{Clinical Alignment}

\subsubsection{Data Generation Pipelines}

\begin{figure}[!ht]
\centering
\includegraphics[width=0.95\linewidth,trim={0.9cm 0cm 0.6cm 0cm},clip]{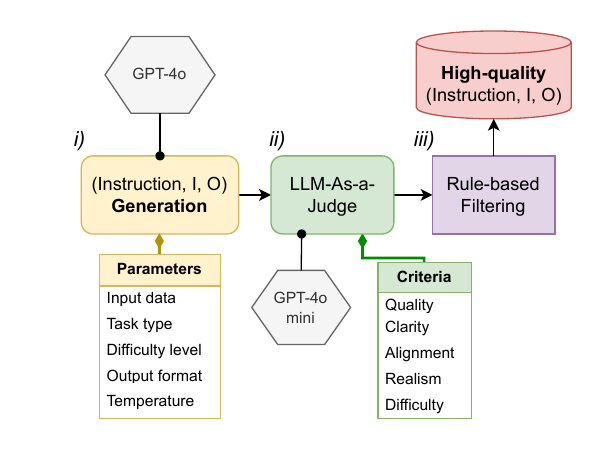}
\caption{Schema of MediFlow generation. i) Given a randomly sampled set of parameter values, we prompt \textit{GPT-4o} for $N$ instructions to obtain triplets $(instruction, inputs, outputs)$ in which we have $P$ pairs of inputs ($I$) and outputs ($O$) each. ii) We prompt \textit{GPT-4o mini} with LLM-as-a-Judge and self-consistency on $M$ samples. iii) We define a heuristic to filter a diverse, high-quality subset for alignment purposes.}
\label{fig:mediflow_diagram}
\end{figure}

\paragraph{Generation of MediFlow}
In Figure \ref{fig:mediflow_diagram}, we illustrate our agentic pipeline to generate the MediFlow dataset.

At step \textit{i)}, we prompt \textit{GPT-4o} with a meta-prompt to generate several triplets $(instruction, inputs, outputs)$ at once conditioned on five parameters: input-data type, task type, difficulty level, output format, and temperature. For this step, we also apply temperatures of 1.0 (70\% of the dataset) or 1.25 (30\% of the dataset), to strike a balance between accuracy or diversity, respectively. Specifically, we request 10 instructions at a time with four input-output pairs each.

In step \textit{ii)}, we prompt \textit{GPT-4o mini} with a LLM-as-a-Judge approach --- using self-consistency with chain-of-thought across $M=5$ samples at temperature of 1.0 --- to provide a critical assessment of the synthetic instruction. We provide five criteria in the prompt on a scale of 1 to 4: quality, clarity, alignment, realism, and difficulty. We compute the final score $S_j \in [1, 4]$ of the $j^{th}$ criterion by summing individual scores $s_{ij} \in \{1,2,3,4\}$ multiplied by their respective counts $c_i \in [0, M]$ (constrained by $\sum_{i=1}^M c_i = M$) as in $S_j= \frac{1}{M} \sum_{i=1}^M c_i \cdot s_{ij}$. 

At the final step \textit{iii)}, we use a heuristic to trim the collection down to its top-$K$ highest quality samples based on the quality criteria. We added details of the process in Appendix \ref{sec:judge}.

\paragraph{Generation of MediFlow-DPO}
From MediFlow, we filter the 130k top-quality samples stratified by task type, input-data type and output format to further align the SLM with direct preference optimization (DPO) after SFT. We generate marginally wrong outputs (i.e. rejected outputs) prompting GPT-4o as an error inducer. We provide the detail in Appendix \ref{sec:appendix_dpo}.

\subsubsection{Alignment Process SFT + DPO}
We train $MediPhi$, the multi-expert merged SLM, with supervised fine-tuning (SFT) on MediFlow to obtain $MediPhi$-$SFT$. We then align $MediPhi$-$SFT$ with DPO which leads to $MediPhi$-$Instruct$ (SFT + DPO). We provide the hyperparameters for both settings in Tables \ref{tab:sft_params} and \ref{tab:dpo_params} of Appendix \ref{sec:appendix_dapt}.

\section{Experiments}
\subsection{CLUE+ Benchmark}
The CLUE Benchmark \cite{dada2024clue} covers six datasets: MedNLI \cite{shivade2017mednli}, MeQSum \cite{MeQSum}, Problem List Summarization \cite{gao2023bionlp}, LongHealth \cite{adams2024longhealth}, MeDiSumQA \citep{dada2025medisumqa}, and MeDiSumCode. For these datasets, we implement the same configuration (i.e. prompts, metrics, and few shots) as \citet{dada2024clue}. While CLUE focused on six tasks using clinical notes and discharge summaries as input data, we extend it with additional clinical input documents (e.g., radiology reports, doctor-patient dialog) and tasks (e.g. information extraction, medical error detection) to obtain a broader assessment of clinical abilities. We introduce the CLUE+ benchmark, with six additional datasets: MedicationQA \cite{BenAbacha2019medicationqa}, MEDIQA-RRS QA \cite{abacha2021mediqa}, MEDEC \cite{abacha2024medec}, ACI-Bench \cite{yim2023aci}, Social Determinant of Health \cite{lybarger2023n2c2}, and
MedConceptsQA ICD10CM \cite{shoham2024medconceptsqa}. We provide further information in the Appendix \ref{sec:CLUE+} in Tables \ref{tab:cluep_detail} and \ref{tab:cluep_benchmark}.

\subsection{Continual Pre-Training}
To assess the efficiency of different continual pre-training methods, we focus on ICD10CM medical coding webpages from the \emph{MedCode} dataset. After continual pre-training on this subset, we test the models on ICD10CM questions from the MedConceptsQA dataset~\cite{shoham2024medconceptsqa}. We left out the easy and medium difficulty questions to focus on the most challenging questions.

\subsubsection{DAPT vs. Explainer vs. PIT}
\label{sec:cp_methods}

\begin{figure}[h!]
\centering
\includegraphics[width=0.9\linewidth]{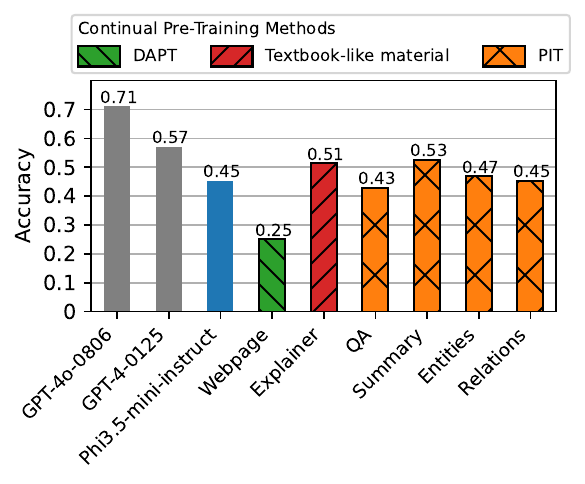}
\caption{Performance of different continual pre-training methods: domain adaptation pre-training (DAPT) on ICD10CM webpages, fine-tuning on synthetic textbook-like material generated with GPT-4o (Explainer), and pre-instruction tuning (PIT) with different synthetically generated tasks: QA, summarization, NER, and relation extraction.}
\label{fig:med_code_exp}
\end{figure}

In Figure \ref{fig:med_code_exp}, we plot the accuracy of different continual pre-training methods DAPT, Explainer and PIT, as described in Sec.~\ref{sec:desc-cp}. First, we note that DAPT dimishes the performance to random ($Webpage$). We hypothesize that the webpages have peculiar implicit format. One solution is to reformulate them as textbook-like explainers. The model fine-tuned on explainers is improving upon the baseline by 6\% ($Explainers$). We see that using PIT by generating summaries ($Summary$) increases performance further by 8\% percent.

\subsubsection{Domain-Specific Merging}
\label{sec:merging_back}

\begin{figure}[h!]
\centering
\includegraphics[width=\linewidth]{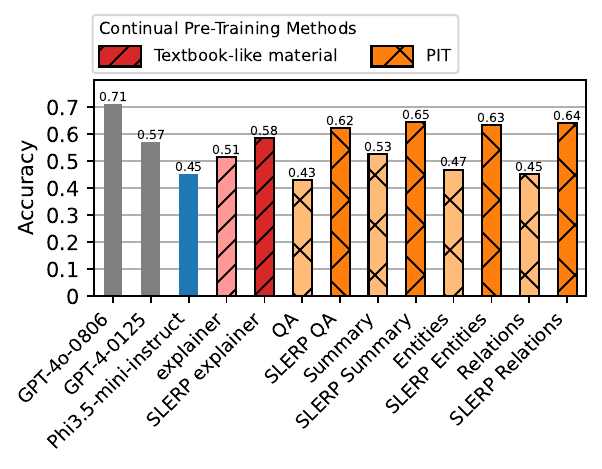}
\caption{Impact of SLERP merging on the performance of ICD10CM. Merging back with the base model (SLERP 50\%) systematically results in gains and these are more pronounced for PIT.}
\label{fig:med_code_pit}
\vspace{-0.3cm}
\end{figure}

In Figure \ref{fig:med_code_pit}, we experiment with merging back into the base model after fine-tuning on the explainers or training with PIT. 

We observe significant improvements across tasks for SLERP-merged models, with the Summary model performing best. Although question answering ($QA$), named entity recognition ($Entities$), and relation extraction ($Relations$) initially decline compared to the base model before merging, SLERP not only boosts the explainer model to 58\% (a 28\% relative improvement over the base model) but also enhances all PIT-trained models further—reaching 62\% (38\% relative) for QA and 65\% (44\% relative) for Summary, surpassing GPT-4 by 8\% (14\% relative).
Based on these results, the rest of the paper will apply PIT with summaries, unless otherwise stated.

\subsubsection{Multi-Domain Merging}
\label{sec:multi_model_merging}

We present the results on the CLUE+ benchmark of the five medical and clinical experts, adapted on the dataset groups and merged back with the base model with SLERP in Table \ref{tab:individual_experts}. While the \textit{DataMix} merged model trained on all the corpora (i.e. similar to BioMistral) achieves gains over the base model, the individual experts realize further gains, except for \textit{MedCode}. The latter improves over the base models but mostly only on coding datasets (i.e. MeDiSumCode and MedConceptsQA, see Tables \ref{tab:clue_detail_results} and \ref{tab:cluep_detail_results} in Appendix \ref{sec:CLUE+} for detailed benchmark). The highest enhancement comes from \textit{MedWiki} on 10 out of 12 datasets, with an average improvement of 3.2\% (8.8\% relative). One possible explanation is that the model is learning better on educational contents --- such as encyclopedic material --- with vast coverage of medical concepts. We observe that removing PIT and/or merging on the \textit{Guideline} group lead to the worst outcomes below the baselines.

\begin{table}[h!]
    \newcommand{\green}[1]{\textcolor[HTML]{008000}{#1}}
    \newcommand{\red}[1]{\textcolor{red}{#1}}
    \caption{Average performances on CLUE+ of SLERP-merged experts obtained with PIT on the five dataset groups. We provide \textit{DataMix} as a baseline SLERP expert trained on all dataset groups. We also provide an ablation study on \textit{Guideline}.{\small We indicate \green{gains} and \red{losses} over the base model. \#DG stands for datasets with gains (out of 12 datasets). CV $\Delta$ stands for coefficient of variations of gains/losses.}}
    \centering
    \renewcommand{\arraystretch}{1.2}
    \begin{tabular}{clccc}
        \toprule
         & & {\scriptsize \textbf{AVG}$\uparrow$} & {\scriptsize \textbf{\#DG}$\uparrow$} & {\scriptsize \textbf{CV $\Delta$}$\downarrow$} \\
         \midrule
        {\small \textbf{Baselines}} & {\small \textit{Phi-3.5 mini}} & 36.5 & - & - \\
        \hline
         \multirow{6.5}{*}{\small \shortstack[c]{\textbf{SLERP}\\\textbf{Experts}}} & {\small \textit{DataMix}} & \green{37.5} & \textbf{10} & \textbf{1.2} \\
         \cmidrule{2-5}
         & {\small \textit{PubMed}} & \green{37.7} & 9 & 1.8 \\  
         & {\small \textit{Clinical}} & \green{39.6} & \textbf{10} & 2.0 \\  
         & {\small \textit{MedWiki}} & \green{\textbf{39.7}} & \textbf{10} & 1.5 \\  
         & {\small \textit{MedCode}} & \green{36.7} & 5 & 39.6 \\  
         & {\small \textit{Guideline}} & \green{39.2} & \textbf{10} & 1.8 \\
         \hline\hline
         \multirow{3.5}{*}{\small \shortstack[c]{Guideline\\Ablations}} & {\scriptsize \ \ \ w/o SLERP} & {\small \red{27.2}} & {\small 4} & {\small 1.0} \\  
         & {\scriptsize \ \ w/o PIT} & {\small \red{33.0}} & {\small 6} & {\small 0.9} \\  
         & {\scriptsize \ \ w/o SLERP\&PIT} & {\small \red{25.2}} & {\small 1} & {\small 2.8} \\
         \bottomrule
    \end{tabular}
    \label{tab:individual_experts}
\end{table}

We present the results on CLUE+ of combining all experts using three multi-model merging techniques (Task-Arithmetic, Ties-merging, and BreadCrumbs) in Table \ref{tab:global_merge}. While Task-Arithmetic yields the highest score of 39.4 with gains on 9 out of 12 benchmark datasets, the models from Ties and BreadCrumbs remain competitive, demonstrating more uniform gains across the benchmark, as indicated by their lower CV $\Delta$. Since the goal is to select the most robust model for the alignment phase, we find that the \textit{BreadCrumbs} expert offers the best trade-off between high-amplitude improvements and consistent gains on 11 datasets. We also see that all unified SLMs are above the average expectation of the SLERP performances by 0.9\%, but below the maximum values and below the top-performing \textit{MedWiki} expert by 0.3\%. Yet, this expert improves only on 10 datasets, one less than the BreadCrumbs-merged expert. We also observe for \textit{BreadCrumbs} a stronger improvement over \textit{MedWiki} of 10.6\% on ICD10CM medical coding (see Table \ref{tab:cluep_detail_results} in Appendix \ref{sec:CLUE+}), which is still lower than the specialized \textit{MedCode} expert with an improvement of 36.9\% over the \textit{MedWiki} expert. We select the \textit{BreadCrumbs} expert as $MediPhi$ for its strong, balanced average score on CLUE+.

\begin{table}[t]
    \newcommand{\green}[1]{\textcolor[HTML]{008000}{#1}}
    \newcommand{\red}[1]{\textcolor{red}{#1}}
    \caption{Average performance on CLUE+ of unifying experts into one SLM. We leverage Task-Arithmetic (TA), TIES-merging and Breadcrumbs (BC) for merging 5 experts: \textit{PubMed}, \textit{Clinical}, \textit{MedWiki}, \textit{MedCode} and \textit{Guideline}. We provide statistics across dataset performances of the SLERP experts, translating into worst case (minimum), average, and best case (maximum).{\small  We indicate \green{gains} and \red{losses} over the baseline. \#DG stands for datasets with gains (out of 12 datasets). CV $\Delta$ stands for coefficient of variations of gains/losses.}}
    \centering
    \renewcommand{\arraystretch}{1.2}
    \begin{tabular}{clccc}
        \toprule
        & & {\scriptsize \textbf{AVG}$\uparrow$} & {\scriptsize \textbf{\#DG}$\uparrow$} & {\scriptsize \textbf{CV $\Delta$} $\downarrow$} \\
        \midrule
        {\small \textbf{Baseline}} & {\small \textit{Phi-3.5 mini}} & 36.5 & - & - \\
        \cmidrule{1-5}
        \multirow{4}{*}{\small \shortstack[c]{SLERP\\Experts}} & {\small Minimum} & \red{34.5} & 4 & 1.9 \\  
        & {\small Average} & \green{38.5} & 7 & 2.1 \\  
        & {\small Maximum} & \green{43.1} & 11 & 1.1 \\
        \cmidrule{2-5}
         & {\small \textit{DataMix}} & \green{37.5} & 10 & \textbf{1.2} \\
        \cmidrule{1-5}
        \multirow{3}{*}{\shortstack[c]{\small \textbf{Unified}\\\textbf{SLM}}} & {\small Task-Arithmetic} & \green{\textbf{39.4}} & 9 & 1.9\\
        & {\small Ties} & \green{39.3} & 7 & 1.7 \\
        & {\small BreadCrumbs} & \green{39.3} & \textbf{11} & 1.5 \\
        \bottomrule
    \end{tabular}
    \label{tab:global_merge}
\end{table}

\subsection{Post-Training}

\begin{table}[ht!]
\newcommand{\green}[1]{\textcolor[HTML]{008000}{#1}}
\newcommand{\red}[1]{\textcolor{red}{#1}}
\caption{Average performance on CLUE+ of aligning Mediphi with MediFlow using SFT followed by DPO.{\small  We indicate \green{gains} and \red{losses} over the baseline. \#DG stands for datasets with gains (out of 12 datasets). CV $\Delta$ stands for coefficient of variations of gains/losses.}}
\centering
\renewcommand{\arraystretch}{1.1}
\begin{tabular}{cllcc}
    \toprule
    & & {\scriptsize \textbf{AVG}$\uparrow$} & {\scriptsize \textbf{\#DG}$\uparrow$} & {\scriptsize \textbf{CV $\Delta$} $\downarrow$} \\
    \midrule
    \multirow{7}{*}{\small \shortstack[c]{Alignment\\of\\SLMs}} & \textit{Phi3.5 mini} & 36.5 & - & - \\
    & \ \ \ +SFT 800K & \green{42.2} & 9 & 1.4 \\
    &  \ \ \ \ \ \ \ +DPO & \green{42.2} & 8 & 1.4 \\
    \cmidrule{2-5}
    & \textit{MediPhi} & \green{39.3} & \textbf{11} & 1.5 \\
    & \ \ \ +SFT 2.5M & \green{41.9} & 9 & 1.6\\
    & \ \ \ +SFT 800K & \green{43.0} & 9 & 1.4 \\
    &  \ \ \ \ \ \ \ +DPO & \green{\textbf{43.4}} & 9 & 1.4 \\
    \bottomrule
\end{tabular}
\label{tab:alignment}
    \vspace{-0.3cm}
\end{table}

We show the results of aligning SLMs with the Mediflow dataset, as well as MediFlow-DPO in Table \ref{tab:alignment}. To begin, we note that the alignment of MediPhi on all instructions (2.5M) leads to an accuracy of 41.9, an improvement of 5.4\% over Phi3.5. By filtering MediFlow for top-quality 800k, we push the gain further by 6.5\% to 43.0\%. Then, we apply DPO using MediFlow-DPO to realize an improvement of 6.9\% (18.9\% relative), of which the maximum average ends up at 43.4\%. Notable gains in relative are: medical entities (SDoH) with 64.3\% and on radiology reports (RRS QA) by 49.5\% (see Table \ref{tab:cluep_detail_results}). If we align Phi3.5 on MediFlow using SFT, we reach an accuracy of 42.2\%. While surpassing $MediPhi$-$SFT$ 2.5M by 0.3\%, $MediPhi$-$SFT$ 800k surpasses it by a margin of 2\% (relative). Despite achieving gains overall, we observe a diminution in $\#DG$ from 11 to 9 for all models. The lower-performing datasets for all aligned models are Problem List Summarization and MediSumCode, which we hypothesize this result from a specific bias in MediFlow towards listing tasks like extracting problems from clinical notes or medical codes from discharge summaries (see Tables \ref{tab:clue_detail_results} and \ref{tab:cluep_detail_results} in Appendix \ref{sec:CLUE+}).

\begin{table}[hb!]
\centering
\setlength{\tabcolsep}{2pt}
\renewcommand{\arraystretch}{1.3}
\newcommand{\green}[1]{\textcolor[HTML]{008000}{#1}}
\newcommand{\red}[1]{\textcolor{red}{#1}}
\caption{Performances on CLUE+ of other medical LLMs compared with MediPhi models.{\small  We indicate \green{gains} and \red{losses}. \#DG stands for datasets with gains (out of 12 datasets). CV $\Delta$ stands for coefficient of variations of gains/losses.}}
\begin{tabular}{llcc}
\toprule
\textbf{Model} & \textbf{\scriptsize AVG $\uparrow$} & {\scriptsize \textbf{\#DG}$\uparrow$} & {\scriptsize \textbf{CV $\Delta$} $\downarrow$} \\
\midrule
Mistral-7B-Instruct-v0.1 & 33.6 & - & - \\
BioMistral-7B-DARE & \green{34.7}  {\small (+1.1)} & 8 & 3.4 \\
\cmidrule{1-4}
Phi-3.5 mini (3.8B) & 36.5 & - & -  \\
MediPhi (3.8B) & \green{39.3}  {\small (+2.8)} & \textbf{11} & 1.5\\
MediPhi-SFT (3.8B) & \green{43.0}  {\small (+6.5)} & 9 & \textbf{1.4} \\
MediPhi-Instruct (3.8B) & \green{\textbf{43.4}}  {\small (+6.9)} & 9 & \textbf{1.4} \\
\cmidrule{1-4}
Meta-Llama-3-8B-Instruct &  44.1 & - & - \\
Llama3-Med42-8B* & \green{45.3} {\small (+1.2)} & 5 & 7.8 \\
\midrule
\multicolumn{4}{l}{\footnotesize *fine-tuned on ACI-Bench, i.e. not a few-shot setting.}
\end{tabular}
\label{tab:clue_results}
\end{table}

\citet{dada2024clue} highlighted a performance gap regarding medical LLMs' performances in clinical settings based on the CLUE benchmark. Out of the twelve medical LLMs in their study, only two are improving upon their base model as shown in Table \ref{tab:clue_results}: BioMistral with DARE model merging, and Med42 with SFT alignment. By applying PIT, merging, and clinical alignment, MediPhi yields the highest improvement over its base model on CLUE+, achieving a +6.9\% gain, compared to +1.1\% for BioMistral and +1.2\% for Med42.

Although Phi3.5 already surpasses Mistral models, the MediPhi SLMs further widen this gap. Despite being less than half the size of LLaMA3 (8B), MediPhi-Instruct (3.8B) achieves near-parity on CLUE+, with a performance difference below 1\%. MediPhi-Instruct outperforms LLaMA3 on four key datasets\footnote{See Tables \ref{tab:clue_detail_results} and \ref{tab:cluep_detail_results} in Appendix \ref{sec:CLUE+}}: ICD10CM (+29.2\%), MeDiSumCode (+13.9\%), RRS QA (+5.8\%), and MeQSum (+3.3\%). Med42 improves over LLaMA3 by +1.2\%, mainly due to a +27.7\% boost on ICD10CM as well as being fine-tuned on the trainset of ACI-Bench. However, MediPhi surpasses Med42 in relative percentages on four tasks: ICD10CM by 2.1\%, on RRS QA by 13.9\%, SDoH by 5.2\%, and MeDiSumCode by 47.6\%. Moreover, it is competitive within 1\% in absolute on three other tasks: MeQSum, MeDiSumQA and MEDEC. Moreover, MediPhi models (3.8B) achieve wider improvements over its base model with a $\#DG$ between 9 and 11, compared to 5 for Med42 (8B). 

\begin{figure}[h!]
\centering
\includegraphics[width=\linewidth]{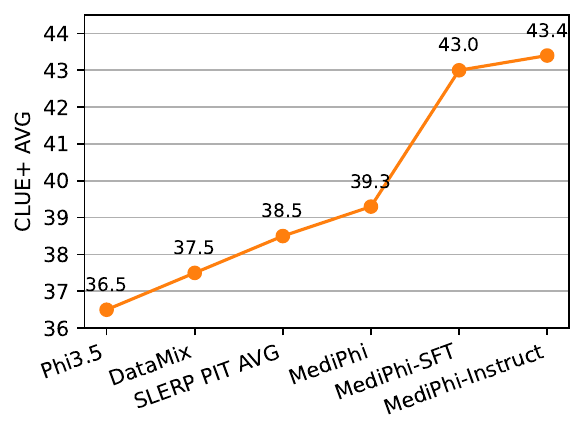}
\caption{Summary of improvements from Phi3.5 to $MediPhi$-$Instruct$ (SFT+DPO). \small{$DataMix$ is adapted on all corpora. $SLERP\ PIT\ AVG$ is the average of five experts trained with PIT on each dataset. $MediPhi$ is the unified expert from the five experts. $MediPhi$-$SFT$ is an instruct model based on SFT only.}}
\label{fig:summary_linechart}
\vspace{-0.5cm}
\end{figure}

We summarized the overall progression of our approach in Figure \ref{fig:summary_linechart}.

\section{Conclusion}

In this work, we introduced MediPhi, the first clinical-focused SLM collection, alongside MediFlow, a large-scale synthetic instruction dataset for clinical alignment. Our results show that PIT significantly enhances domain adaptation, especially when combined with model merging. Notably, our medical coding expert surpasses GPT-4-0125 on the ICD10CM benchmark, and aligning MediPhi with MediFlow improves CLUE+ performance by 18.9\% on average. By releasing these resources, we aim to enhance reproducibility, drive SLM adoption in clinical settings, and foster MediPhi’s continued development through its modular design.

\section*{Limitations}
This study necessitated a vast amount of resources in terms of 8x80GB A100 GPUs on Azure Machine Learning for approximately 12,000 GPU$\cdot$hours, of which close to 3,600 GPU$\cdot$hours were dedicated to achieve the final model with evaluations. It also required access to Azure OpenAI services for GPT-4o, GPT-4o mini and text-embedding-3-large (i.e. close to 25B input-output tokens).

The MediFlow corpus has a few limits. The first limitations is that we set our scope to one-turn instructions instead of multi-turn conversation dataset like UltraChat \cite{ding2023ultrachat}. The second issue is the limited amount of very high complexity tasks (i.e. multi-step reasoning tasks) with also very long input and output. The third limit is that Mediflow is an English-only clinical corpus. To broaden the scope of MediFlow, future works might use seed data from clinical corpora as well as expanding our agentic pipeline. While MediPhi and MediPhi-instruct might conserve abilities from \textit{Phi3.5 mini} (i.e. multilingual, conversational, safety alignment, etc.) by model merging \cite{hammoud2024modelmergesafety}, we hypothesize that these could be affected for which the impact was not studied outside medical and clinical abilities as well as instruction-following abilities. Thus, our recommendation is to use the MediPhi collection specifically on clinical NLP tasks. We also strongly advice a verification of the model output by an expert in the specific medical field of the task.

Our CLUE+ benchmark expands upon the CLUE coverage in terms of tasks, input data and datasets. While the coverage of the clinical field is large, a couple of gaps still remain. The information-extraction tasks are only represented by the SDoH dataset, and there is no input data on the nursing sub-field (e.g. nurse-patient dialogs or notes).

Given the quick evolutions of OpenAI's GPT-4o models as well as their stochastic nature, future exact reproduction of parts of this work may become impossible if the mentioned versions are no longer maintained.

\section*{Ethics Statement}
We acknowledge that the substantial computational resources required for this work, including GPU hours and API calls, contribute to carbon emissions as well as limit the reproduction of this work. However, we believe that the knowledge and artifacts produced --- such as the MediPhi SLM collection and the MediFlow corpus --- offer significant positive impacts. These include enabling the adoption of SLMs in clinical settings, potentially reducing carbon emissions in the medium to long term, and fostering further research on continual pre-training in medical, clinical, and other domains.

Additionally, our modular approach with releases of models and datasets allows researchers and practitioners to build directly on this work, promoting accessibility and collaboration. Strong language models in the clinical field can positively impact public health and support the work of healthcare providers, enhancing patient care and operational efficiency.


\bibliography{custom}
\bibliographystyle{acl_natbib}

\appendix

\section{Appendix}
\label{sec:prompt_appendix}

\subsection{Detail about Dataset Groups}
\label{sec:dapt_datasets}

The \textit{PubMed} group is the largest by 2 orders of magnitude with about 48B tokens. PubMed Central \cite{PMC_Open_Access_Subset} have a segment for commercial use, while abstracts are public.

In the \textit{Medical} group, we have medical Wikipedia known as MedWiki \cite{corbeil2024iryonlp}. We also fetch the open medical guidelines \cite{chen2023meditron} which comes from multiple recognized health organizations (e.g. WHO). Then, we gather medical coding corpora from public websites: ICD9CM, ICD9PROC, ICD10CM, ICD10PROC and ATC. The ICD coding is a wide taxonomy of diseases and medical conditions as well as procedure codes, while \textit{Anatomical Therapeutic Chemical} (ATC) codes are a classification of medical drugs.

In the \textit{Clinical} group, we leverage PMC-Patients v2 \cite{zhao2022pmcpatients} --- subset with distribution- and commercial-friendly CC licenses --- which contains clinical cases. We also include synthetic derivative datasets under same licensing conditions: \textit{Asclepius} \cite{kweon2024asclepius} containing clinical documents and \textit{NoteChat} \cite{wang2023notechat} containing doctor-patient dialogues. Furthermore, we include \textit{MTSamples} \cite{mtsamples}, a public database filled with various de-identified clinical documents.

\subsection{Hyperparameters \& Pre-Training Optimization Strategies}
\label{sec:appendix_dapt}

We use the HuggingFace ecosystem (\textit{transformers}, \textit{trl}, \textit{accelerate}, and \textit{datasets}) to train models leveraging its multi-node setting. The hyperparameters of the continual pre-training are listed in Table \ref{tab:cpt_params}. The hyperparameters of the alignment process are provided in Tables \ref{tab:sft_params} and \ref{tab:dpo_params} for SFT and DPO, respectively.

\begin{table}[!b]
\small
\renewcommand{\arraystretch}{1.3}
\centering
\caption{Hyperparameters for CPT excluding PIT.}
\begin{tabular}{lc}
\toprule
 \textbf{Hyperparameter} &  \\
 \midrule
 Maximum Tokens & 4,096 \\
Optimizer & AdamW \\
LR Scheduler & Linear Warmup - Cosine \\
\# Warmup Steps & 35 \\
Maximum LR & $1\ \times\ 10 ^{-4}$ \\
Epochs & 1 \\
NEFTune $\alpha$ & 5 \\
Batch Size per GPU & 16 \\
\# GPUs & 16 \\
Gradient Accumulation & 2 \\
Effective Batch Size & 512 \\
\bottomrule
\end{tabular}
\label{tab:cpt_params}
\end{table}

\begin{table}[!b]
\small
\renewcommand{\arraystretch}{1.3}
\centering
\caption{Hyperparameters for SFT.}
\begin{tabular}{lc}
\toprule
 \textbf{Hyperparameter} &  \\
 \midrule
Optimizer & AdamW \\
LR Scheduler & Linear Warmup - Cosine \\
\# Warmup Steps & 40 \\
Maximum LR & $2\ \times\ 10 ^{-5}$ \\
Epochs & 2 \\
NEFTune $\alpha$ & 5 \\
Batch Size per GPU & 16 \\
\# GPUs & 8 \\
Gradient Accumulation & 2 \\
Effective Batch Size & 256 \\
\bottomrule
\end{tabular}
\label{tab:sft_params}
\end{table}

\begin{table}[!h]
\small
\renewcommand{\arraystretch}{1.3}
\centering
\caption{Hyperparameters for DPO.}
\begin{tabular}{lc}
\toprule
 \textbf{Hyperparameter} &  \\
 \midrule
Optimizer & AdamW \\
LR Scheduler & Linear Warmup - Cosine \\
\# Warmup Steps & 50 \\
Maximum LR & $1\ \times\ 10 ^{-6}$ \\
Epoch & 1 \\
Batch Size per GPU & 8 \\
\# GPUs & 16 \\
Gradient Accumulation & 1 \\
Effective Batch Size & 128 \\
$\beta$ & 0.1 \\
\bottomrule
\end{tabular}
\label{tab:dpo_params}
\end{table}

\citet{ding2024fewertruncations} highlighted the importance of longer context windows, minimal truncations, and restricted cross-document attention during pre-training. Many prior works in medical NLP rely on concatenate-and-truncate strategies, despite introducing significant truncation \cite{toma2023clinical,med42,christophe2024med42v2,labrak2024biomistral}. Instead, we implement Best-Fit Packing \cite{ding2024fewertruncations} for the \textit{PubMed} expert and \textit{DataMix} baseline, ensuring efficient token utilization. The other experts are trained with PIT one document at a time. For all expert models, we avoid padding and cross-document attention \cite{kundu2024paddingfree} in trainings.

\subsection{Validation Sets}
\label{sec:valsets}

The twelve validation sets cover subjects such as clinical case, clinical knowledge, medication, icd10cm code definitions, radiology report, clinical NLI, QA on discharge letter, medical codes of discharge letter, problem list from clinical notes, summarization of patient inquiry, QA on medical consultation and QA on multiple EHR documents. We control its diversity by generating the embeddings of the question with its context, and applying the density-based clustering method HDBSCAN \cite{mcinnes2017hdbscan}. The final valid sets are the combination of the outliers (i.e. not assigned to a cluster) and one sample per cluster for a total up to 1,200 samples each.

\subsection{Generation of MediFlow triplets}

We parametrized the prompt to generate MediFlow based on the task type, input data, output format (plain text or JSON), difficulty level (moderate, moderate-hard, hard, very hard, or extreme), and number of input-output example pairs (3 or 4 per instruction). For the difficulty level, we favored the sampling of hard, very hard and extreme levels by a ratio of 3:1.

The \textbf{14 task types} are: summarization, question-answering, multiple-choice question-answering, named entity recognition, Relation extraction, classification, reasoning and diagnosis, textual entailment, text simplification, text expansion, abbreviation expansion, aspect-oriented keyword extraction, error detection and correction and note scoring.

The 36 input-data types with various levels of granularity (1 (only complete document) up to 7 (complete document with 6 individual sections)) resulting in a total of 98 fine-grained input-data types. Decisions for the granularity levels were made based on the document length --- e.g. taking the complete short documents compared to segmenting long documents. The data types are:

\begin{enumerate}[noitemsep]
    \item Discharge Summary (6)
    \item SOAP Clinical Note (5)
    \item Clinical Note (6)
    \item Progress Note (4)
    \item Admission Note (6)
    \item Scientific Article (8)
    \item Clinical Case (3)
    \item Nursing Note (1)
    \item Monitoring Data of Vital Signs (1)
    \item Referral Letter (1)
    \item Emergency Department Note (7)
    \item Laboratory Report (1)
    \item Radiology Report (3)
    \item Doctor-Patient Conversation (5)
    \item Nurse-Patient Dialog (5)
    \item Operative Note (8)
    \item Consultation Note (1)
    \item Pathology Report (1)
    \item Prescription Note (1)
    \item Preoperative Assessment (1)
    \item Postoperative Note (1)
    \item Therapy Notes (5)
    \item Immunization Record (1)
    \item Screening Report (1)
    \item Consent Form (1)
    \item Care Plan (1)
    \item Dietary Notes (1)
    \item Psychiatric Evaluation (5)
    \item Social Work Note (1)
    \item End-of-Life Care Documentation (1)
    \item Triage Note (1)
    \item Dental Record (1)
    \item Home Health Care Report (1)
    \item Genetic Testing Report (1)
    \item Incident Report (1)
    \item Patient Education Material (1)
\end{enumerate}

\begin{figure}[!ht]
\centering
\includegraphics[width=0.8\linewidth]{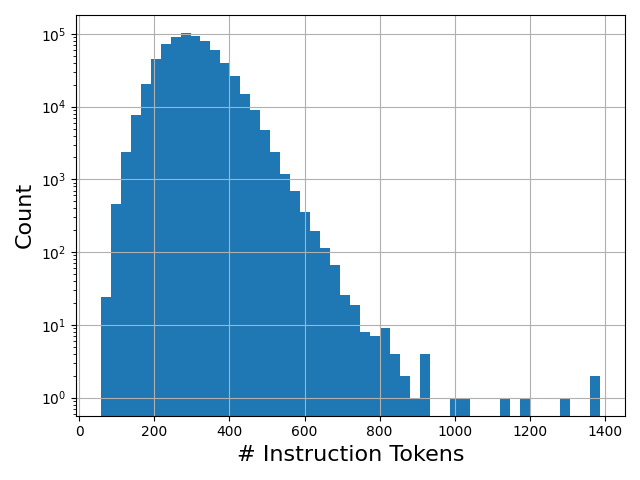}
\caption{Distribution of instruction tokens in MediFlow with y-axis in log scale. The average is $301\pm295$ tokens.}
\label{fig:inst_token_dist}
\end{figure}

\begin{figure}[!ht]
\centering
\includegraphics[width=0.8\linewidth]{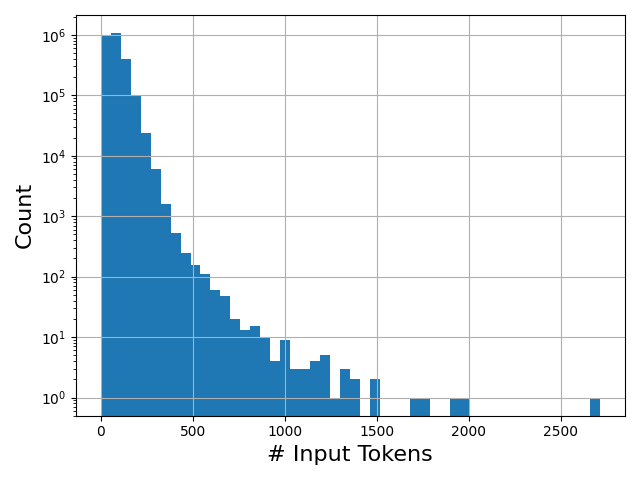}
\caption{Distribution of input tokens in MediFlow with y-axis in log scale. The average is $76\pm67$ tokens.}
\label{fig:input_token_dist}
\end{figure}

\begin{figure}[!ht]
\centering
\includegraphics[width=0.8\linewidth]{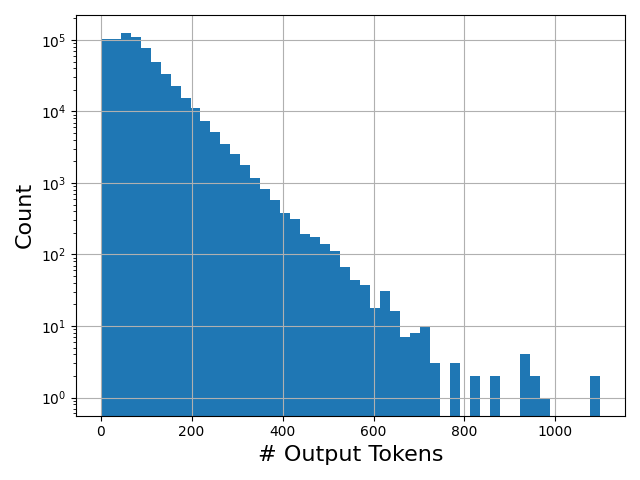}
\caption{Distribution of output tokens in MediFlow with y-axis in log scale. The average is $79\pm66$ tokens.}
\label{fig:output_token_dist}
\end{figure}

In Figures \ref{fig:inst_token_dist}, \ref{fig:input_token_dist} and \ref{fig:output_token_dist}, we present the histograms of tokens for generated instructions, input and output, respectively.

\newpage
\begin{promptbox}[MediFlow Prompt]
\small
You are an expert user querying about the medical and clinical domain in natural language processing.
You will define instructions for a precise task with clear constraints in the medical/clinical domain.
You must be very detailed in the instructions regarding input data, the \{\{task\_type\}\} task and the desired output format which is \{\{output\_format\}\}.
You must use \{\{input\_data\}\} as the task's input in some ways, especially \{\{input\_data\_granular\}\}.
You must put these between the tags <instruction>...</instruction>.
You must define clearly based on these parameters a specific task from the given type, specific expected input data and output format.
You must make a task with a \{\{difficulty\}\} difficulty on a scale of 6 levels (low, moderate, moderate-hard, hard, very hard and extreme). You must not mention the task level in your own instructions.
You must only write the instructions, i.e. do not use markdown, no extra comment, etc.

Then, you must give \{\{number\_examples\}\} examples between tags <examples> <example> <input>...</input> <output>...</output> </example> </examples> containing input and output.
You must give a complete example with input and output at the end.

You must use interesting and complex examples requiring abstractive medical capabilities to infer the output from the input. So, you must avoid any obvious input and output, and you must favor very difficult pairs.
You must strictly use the format <instruction>...</instruction> followed by <examples> <example><input>...</input> <output>...</output> </example></examples> with exactly those tag names.

You must use synonyms for all headers (or no header at all) to avoid leaking current vocabulary into the instructions, also use different ways to structure (or not) and detail the instructions (e.g. bullet points, sections, narrative form, or else).
\end{promptbox}

\subsection{Generation of MediFlow Judge Scores}
\label{sec:judge}
We used GPT-4o-mini to score all MediFlow on 5 criteria: quality, alignment with instruction requirements, coherence, realism and difficulty. We applied a temperature of $1.0$ with chain-of-thought for each criterion and average scores accross 5 generated samples for self-consistency. All scores are an integer on a scale from 1 to 4 as defined below.

\begin{promptbox}[LLM-as-a-Judge MediFlow Prompt]
\small
You are the best instruction designer for language models in the medical/clinical field.
You will be given instructions with constraints for a task to perform on clinical documents.
Your task is to give a critical assessment of the instructions as a nested JSON object. For each criteria as a key, the value is a JSON object containing a "rationale" along a "score" on a scale of 1 to 4. The criteria are: quality, alignment with instruction requirements, coherence, realism and difficulty.

INSTRUCTION REQUIREMENTS:::

Here are the instruction requirements:

- Defined very detailed instructions for a precise task with clear constraints in the medical/clinical domain.

- Clearly defined task type, specific input data and output format. It can also contain examples.

- Only write the instructions, i.e. do not use markdown, no extra comment, etc.

SCALES:::

\{\{criteria\_definitions\}\}

INSTRUCTIONS:::

\{\{instruction\}\}

You must only output the JSON object for the critical assessment. Remember valid scores are: 1, 2, 3 and 4.

OUTPUT:::
\end{promptbox}

\subsubsection{Judge Score Distributions}

\begin{figure}[!ht]
\centering
\includegraphics[width=1\linewidth]{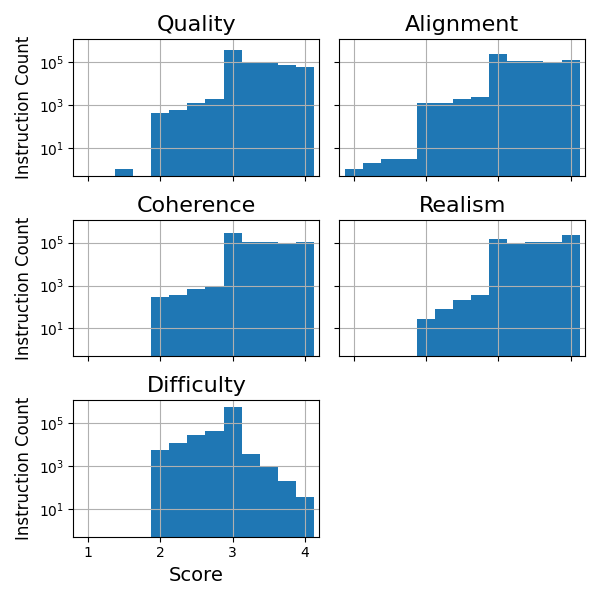}
\caption{Judge Score Distributions across MediFlow. Y-axis is set to log scale.}
\label{fig:judge_all}
\end{figure}

In Figure \ref{fig:judge_all}, we displayed the histograms (with a logarithm y axis) of the judge scores predicted on nearly 700k instructions. For all criteria, we note that a large amount of generated instructions have a peak at 3 with large distribution between 3 and 4 (on a scale from 1 to 4), which is considered on the high end. We observe a different trend with respect to the difficulty criterion, where a larger peak is still at 3 (i.e. difficulty level of hard) but some portion of the dataset (i.e. less than a 100k instruction) have scores between 2 and 3.

\subsection{MediFlow Scatterplots}
\begin{figure*}[!h]
\centering
\includegraphics[width=1.07\linewidth]{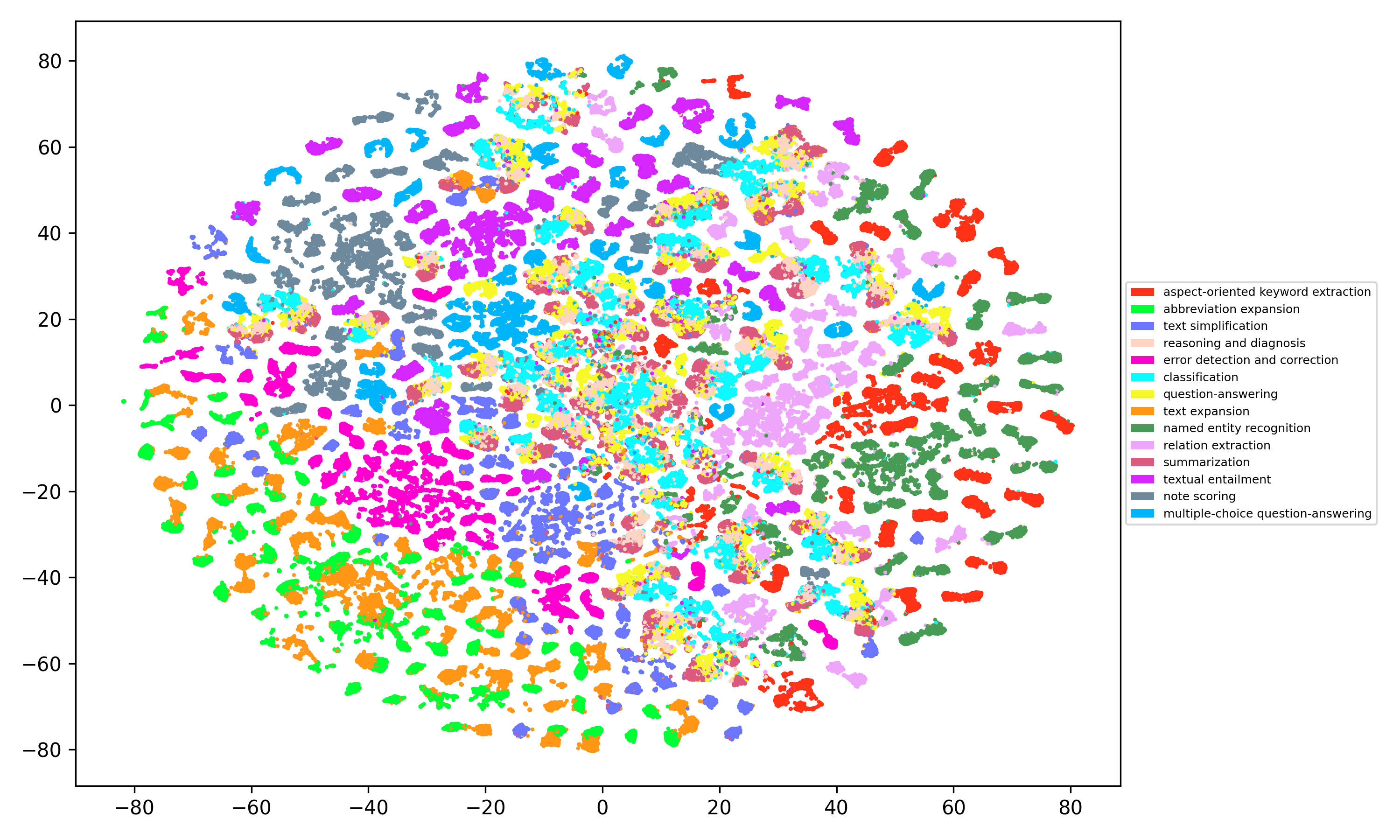}
\caption{t-SNE 2D scatterplot of MediFlow (2.5M) using OpenAI \textit{text-embedding-3 large} API with tasks as colors.}
\label{fig:tsne_mediflow_task}
\end{figure*}

\begin{figure*}[!h]
\centering
\includegraphics[width=1.07\linewidth]{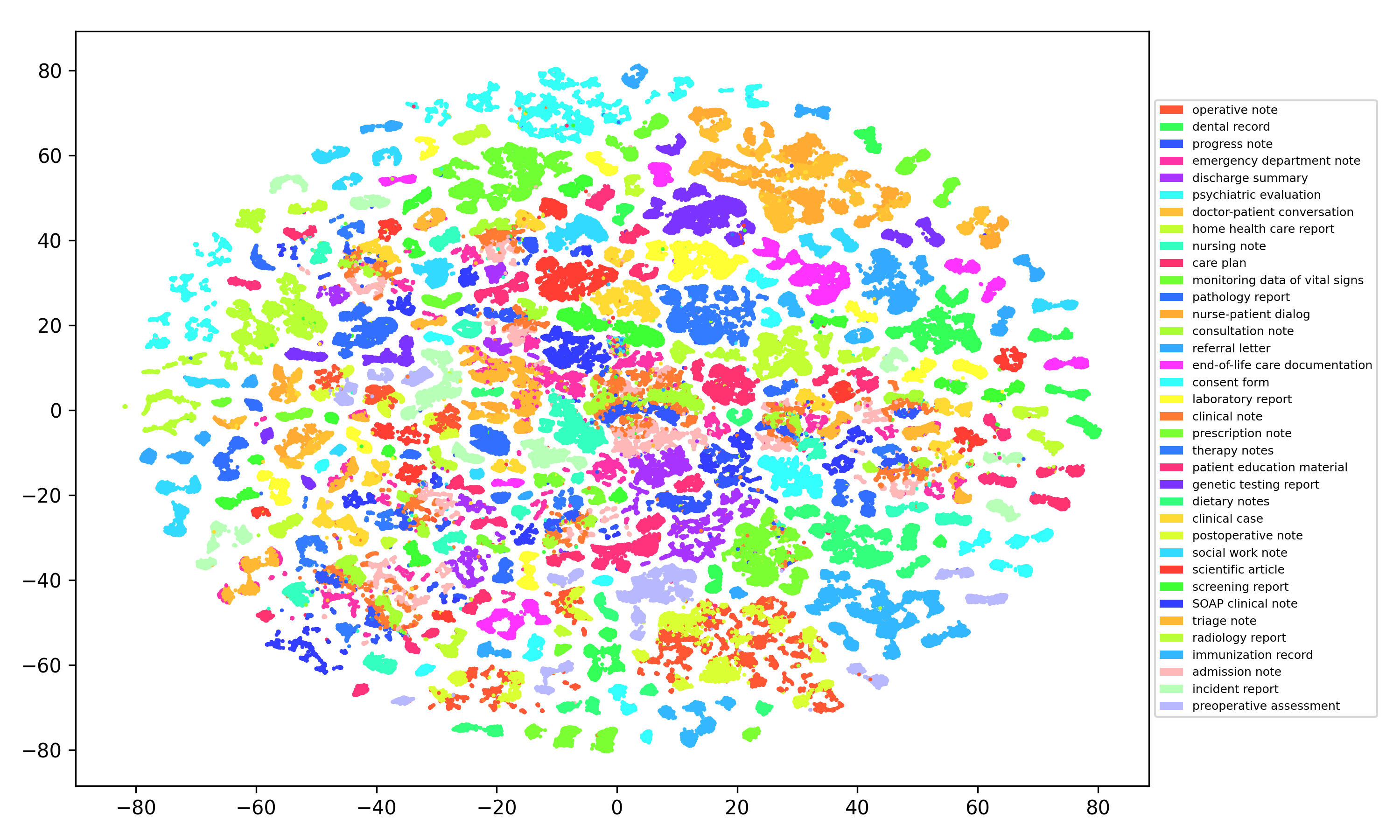}
\caption{t-SNE 2D scatterplot of MediFlow (2.5M) using OpenAI \textit{text-embedding-3 large} API with input-data types as colors.}
\label{fig:tsne_mediflow_input}
\end{figure*}

\begin{figure}[!h]
\centering
\includegraphics[width=\linewidth, trim={0.5cm 0cm 0.3cm 0cm}, clip]{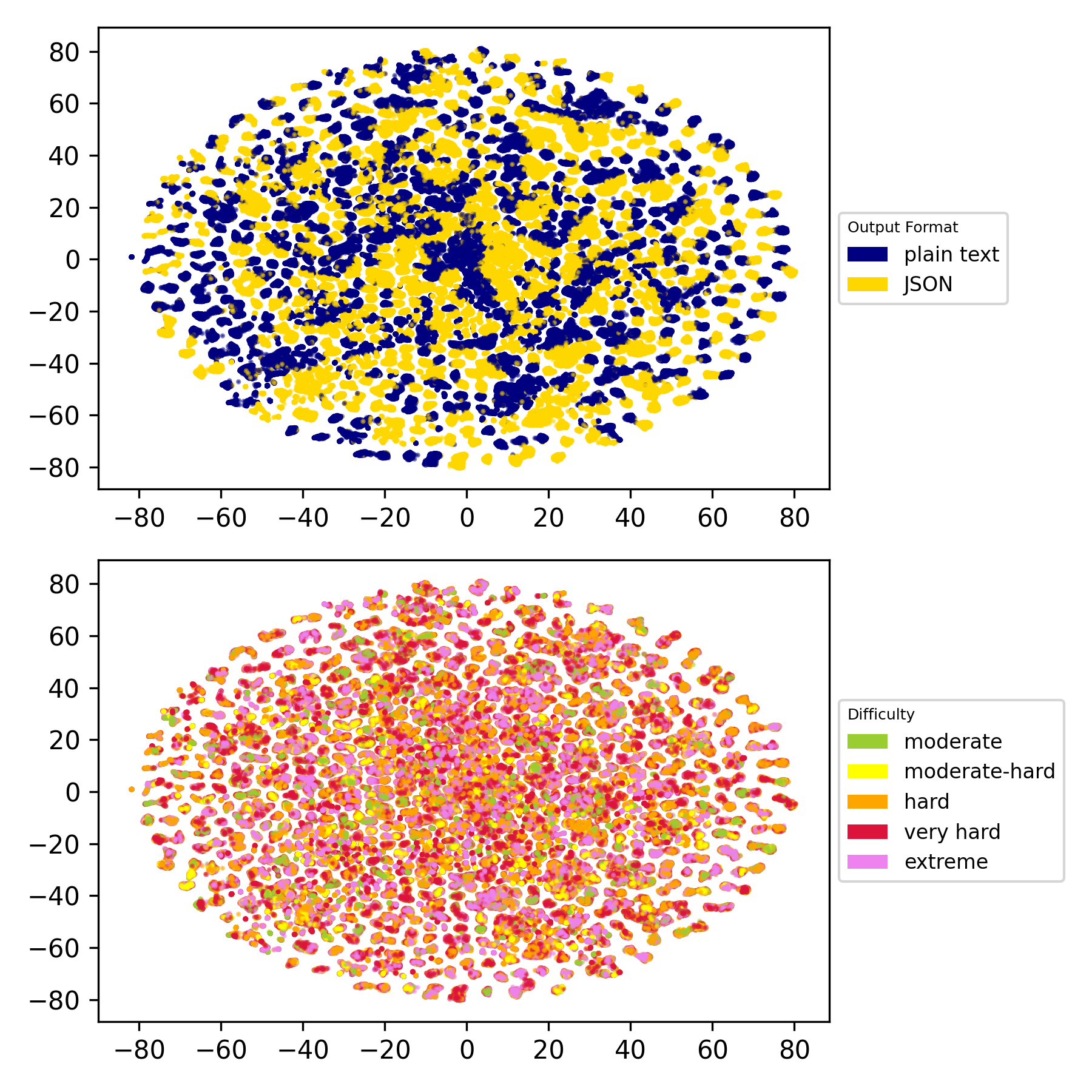}
\caption{t-SNE 2D scatterplots of MediFlow (2.5M) with output format (top) and difficulty level (bottom) as colors.}
\label{fig:tsne_mediflow_output_difficulty}
\end{figure}

We generated embedding with OpenAI \textit{text-embedding-3-large} (truncated at 256 dimensions) for each instruction in MediFlow. Then, we applied PCA at 50 dimensions and t-SNE to get 2 dimensions which we displayed on scatterplots with other labels (output formats, difficulty levels, task type and input-data type) in Figures \ref{fig:tsne_mediflow_output_difficulty}, \ref{fig:tsne_mediflow_task} and \ref{fig:tsne_mediflow_input}. While the scatter plot on the difficulty levels do not exhibit clear clustering patterns, we see that the others have distinctive patterns at different scales. The output format is affecting local clustering patterns often appearing at small scale as two closed blobs. The task and input-data types affect the macrostructure at similar large scale.

\subsection{Generation of Marginally Wrong Output for DPO}
\label{sec:appendix_dpo}
DPO requires to have a prompt (corresponding in our framework to \textit{instruction} with \textit{input}) along a chosen response (i.e. \textit{output}) and a rejected response. The rejected response must be less preferable compared to the chosen one. First, we filter MediFlow to around 85k instructions keeping only the top triplets on the \textit{quality} metric along with a stratification by task type, input-data type and output format. To have the best trade-off between diversity and high quality, we sampled 3 input-output pairs for the first top high-quality 20k instructions, followed by 2 pairs for the next 25k and one pair for the last 40k. After filtering, this resulted the MediPhi-DPO dataset of 130,852 triplets. Then, we prompted GPT-4o with a triplet at a temperature of 1.0 with a randomly sampled error type on the following prompt to generate a marginally wrong output as rejected output. The error types are : ambiguity, partial correctness, over-verbosity, brevity, unbalanced detail, stylistic issues, factual inaccuracy, logical flaws, misinterpretation, simplistic reasoning, grammatical errors, and spelling errors.

\begin{promptbox}[Marginally Worst Output Prompt]
\small
You are a subtle flaw introducer, trained to degrade a high-quality response by introducing a specific type of error without making it obviously incorrect. You will receive the triplet (instructions, input, output). The instructions are what to do to the input to get the output, which is the good response. Your task is to provide a wrong output but in a subtle way. Here's the definitions of error types:

\{\{error\_type\_definitions\}\}

INSTRUCTIONS
\{\{instruction\}\}

INPUT
\{\{input\}\}

OUTPUT
\{\{output\}\}

Generate a wrong response that is marginally worse than the good output. Introduce an error of type \{\{error\_type\}\}, ensuring the response still seems reasonable at first glance.
\end{promptbox}

\subsection{CLUE+ Benchmark}
\label{sec:CLUE+}

\subsubsection{MedicationQA}

MedicationQA \cite{BenAbacha2019medicationqa} consists of $674$ consumer health questions collected from MedlinePlus\footnote{https://medlineplus.gov}. These questions were linked to a matching excerpt from a trusted source that contains the answer.

\begin{figure*}[ht!]
\centering
\includegraphics[width=1.05\textwidth]{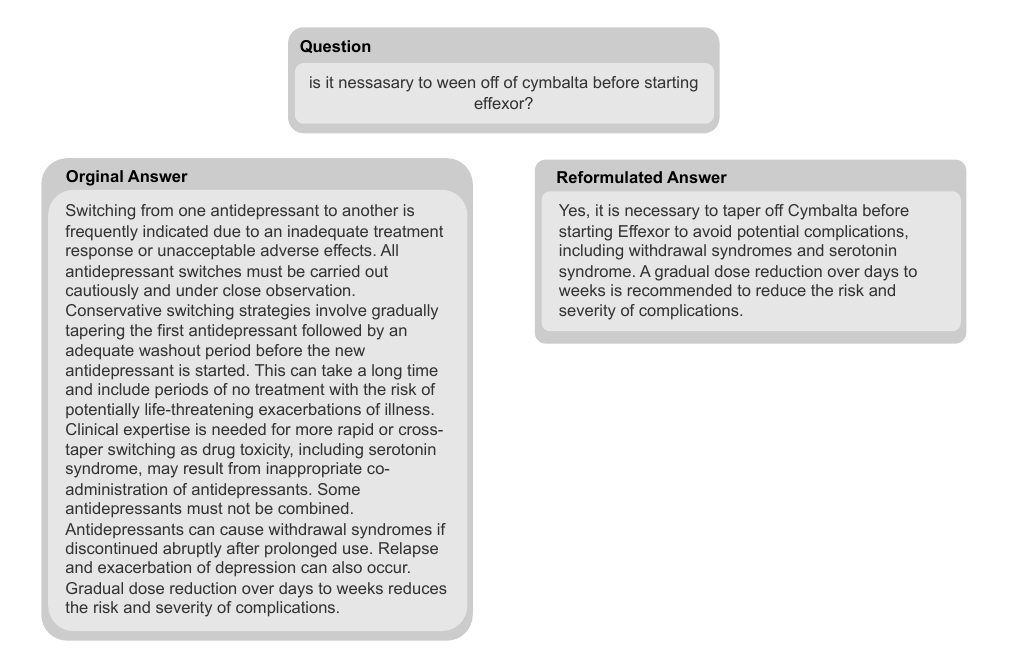}
\caption{An example of a verbose answer in MedicationQA and the reformulated answer we replaced it with.}
\label{fig:medqa_reformulation}
\end{figure*}

Through manual inspection we identified two obstacles for effective few-shot evaluation of LLMs on MedicationQA:

\begin{itemize}[noitemsep]
    \item Some questions were poorly formulated or a search query instead of a question
    \item Some of the answers did not give a specific answer to the question
    \item Answers were often unconcise since they were not formulated as a direct answer to the question, but a retrieved excerpt from a text
\end{itemize}
Based on these observations we prompted an LLM\footnote{\label{fn:llm}meta-llama/Llama-3.3-70B-Instruct} to remove non-matching question answer pairs and formulate a direct answer based on the given excerpt. Figure \ref{fig:medqa_reformulation} shows an example for a reformulated answer. This process resulted in $485$ question-answer pairs. 

For benchmarking we use the following system prompt:

\begin{promptbox}[MedicationQA System Prompt]
\label{medicationQAPrompt}
\scriptsize
You are a highly skilled assistant, specifically trained to assist patients with medical questions. Give a concise answer. Do not mention anything that was not explicitly asked for. Do not generate anything else.
\end{promptbox}

\begin{figure*}[ht!]
\centering
\includegraphics[width=1.05\textwidth]{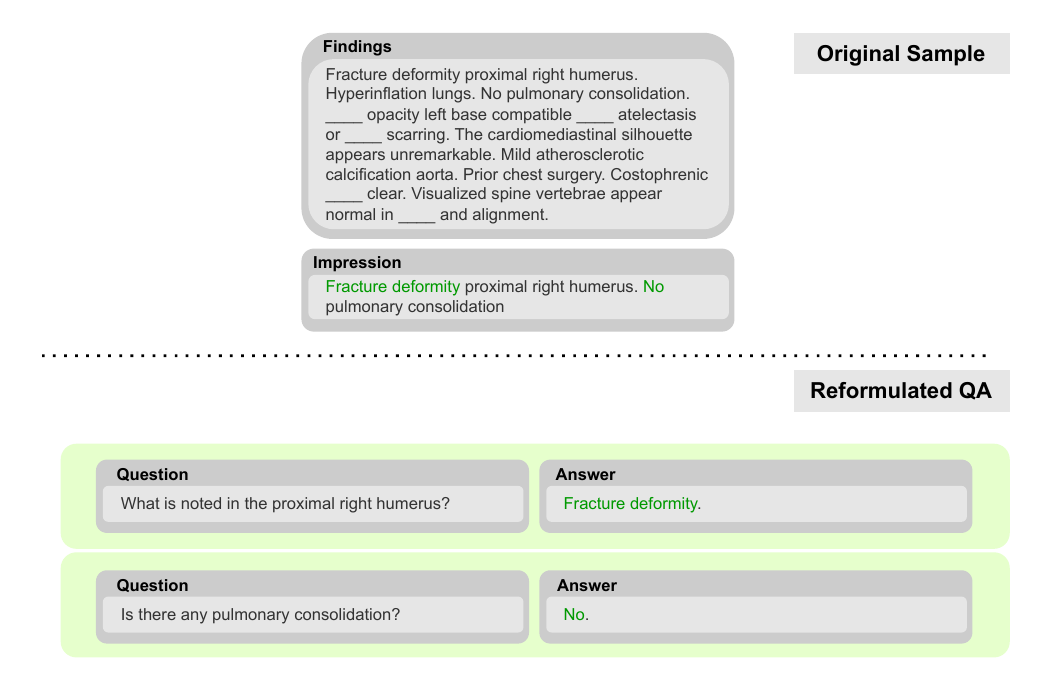}
\caption{An example of a formulation of questions based on the impression section.}
\label{fig:rrs_reformulation}
\end{figure*}

\subsubsection{MEDIQA-RRS QA}
MeDIQA-RRS \citep{abacha2021mediqa} is a summarization dataset based on findings and impressions sections of Radiology reports from the Indiana University chest X-ray dataset \citep{demner2016preparing} and reports from the Stanford Health Care system. The findings serve as model inputs while the impressions are treated as the summarization ground truth. We evaluate on the test split, which contains 600 finding-impression pairs. We observed that the impressions were often not a complete summary, but rather an answer to a question posed before the exam (e.g., "No acute cardiopulmonary findings."). Additionally, the comprehensiveness of impressions varied substantially between reports. To address this we reformulated the impressions to a series of question-answer pairs using an LLM\footref{fn:llm}. We specified that the answers should use the same wording as in the original impression section to ensure factuality. We confirmed that answers were not changed, filtering answers with exact string matching. Figure \ref{fig:rrs_reformulation} shows an example of this reformulation step.

The following system prompt was used in the evaluation:

\begin{promptbox}[MEDIQA-RRS QA System Prompt]
\scriptsize
You are a highly skilled assistant, specifically trained to interpret radiology reports. You will receive the findings section of a report along with specific questions. Provide concise, focused answers based solely on the information provided. Avoid adding any details not explicitly requested.
\end{promptbox}

\subsubsection{ACI-Bench}
ACI-Bench \cite{yim2023aci} is a dataset that takes as input a doctor-patient dialog and the task is to generate a clinical note of five sections.

\begin{promptbox}[ACI-Bench System Prompt]
\scriptsize
Task: Generate an Extremely Detailed Clinical Note from a Doctor-Patient Conversation

Role: You are an expert medical professional responsible for creating a highly detailed, comprehensive, and fully elaborated clinical note from a doctor-patient conversation.

Instructions:
- You will receive a doctor-patient conversation as input.
- Your task is to produce a long, exhaustive clinical note covering all clinically relevant details.
- The note must be extremely detailed, ensuring no important information is omitted.
- Expand on every symptom, examination finding, and treatment plan to provide a complete, structured summary.

Output Format:
Your clinical note must be structured into the following five required sections:

1. CHIEF COMPLAINT
A precise and natural-language statement describing the primary reason for the visit. Usually written in the patient’s own words (e.g., "Chest pain for two days.").

2. HISTORY OF PRESENT ILLNESS
Provide a detailed, full narrative including:
- Onset: Exact time course (sudden/gradual, exact duration).
- Duration: Progression over time.
- Severity: Patient’s description or numeric scale (1-10).
- Location: Anatomical specificity.
- Modifying Factors: What worsens or improves symptoms (activities, medications).
- Associated Symptoms: Describe all related symptoms.
- Prior Treatments: List exact medications, dosages, patient responses.

3. PHYSICAL EXAM
Expand on every finding instead of using short labels.
Describe specific observations for each system:
- Vital Signs: BP, HR, Temp, O2 Sat, RR.
- General Appearance: Patient’s demeanor, level of distress.
- Neurological: Reflexes, motor strength, sensory findings.
- Cardiovascular: Detailed heart sounds, pulses, peripheral findings.
- Pulmonary: Breath sounds, presence of wheezes/crackles.
- Abdominal: Bowel sounds, tenderness, distension.

4. RESULTS
Include all relevant diagnostic data, explaining why each result matters. List both abnormal and pertinent normal findings.

5. ASSESSMENT AND PLAN (A/P)
Diagnosis \& Differential Diagnosis: Explain why the most likely condition was chosen.
Plan:
- Medications: Name, dose, frequency, and rationale.
- Additional Tests: Imaging, lab work, specialist referrals.
- Follow-up Plan: Next steps, expected outcomes.
- Patient Education: Instructions, lifestyle modifications.
- Ensure that every plan component is justified.

Reminder:

- You must include every detail exhaustively from the dialog.

- You must justify every diagnosis and treatment.

- You must provide a thorough explanation of the assessment and plan.

Thus, you must expand on every detail from the dialog in the note.
\end{promptbox}

\subsubsection{Social Determinants of Health}

\begin{promptbox}[SDoH System Prompt]
\scriptsize
Task: Entity Extraction for Social Determinants of Health (SDOH)
Your job is to extract key socio-demographic and behavioral factors from the provided text. Your output must be a list of JSON objects, where each object contains:

- "entity": The specific category of the extracted information, chosen from the predefined taxonomy below.
- "label": The corresponding label from the taxonomy.
- "value": The exact phrase from the document that represents the entity.

Entities \& Taxonomy:

1. Employment

- "entity": "Employment" (general employment-related mention)

    - "label": "StatusEmploy" → "employed", "unemployed", "retired", "on disability", "student", "homemaker"
    
    - "label": "Duration" → "for the last five years", "since 2010"
    
    - "label": "History" → "15 years ago", "in 2005"
    
    - "label": "Type" → Specific occupations (e.g., "geologist", "registered nurse", "office work")

2. Living Status

-"entity": "LivingStatus" (mentions of where and how someone lives)

    - "label": "StatusTime" → "current", "past", "future"
    
    - "label": "TypeLiving" → "alone", "with family", "with others", "homeless"
    
    - "label": "Duration" → "for the past ten years", "since 2015"
    
    - "label": "History" → "moved out five years ago", "in 2010"

3. Substance Use

- "entity": "Alcohol", "Drug", "Tobacco" (mentions of substance use)

    - "label": "StatusTime" → "none", "current", "past"
    
    - "label": "Duration" → "for the past eight years"
    
    - "label": "History" → "seven years ago", "in 2005"
    
    - "label": "Method" → "smoke", "snort", "inhale", "inject" (for drugs), "chew", "vape" (for tobacco)
    
    - "label": "Type" → "beer", "wine", "heroin", "marijuana", "cigarettes"
    
    - "label": "Amount" → "\# of drinks", "\# of cigarettes", "\# of times"
    
    - "label": "Frequency" → "daily", "monthly", "yearly"

Example Output (Using Entities from the Document):
[
  {"entity": "Employment", "label": "StatusEmploy", "value": "full-time student"},
  {"entity": "LivingStatus", "label": "TypeLiving", "value": "currently lives alone"},
  {"entity": "Alcohol", "label": "StatusTime", "value": "drinks occasionally"},
  {"entity": "Drug", "label": "History", "value": "used marijuana seven years ago"}
]

Guidelines for Extraction:

1. Extract only explicitly mentioned entities—do not infer information.

2. Use exact text from the document—the "value" must match the original wording.

3. Categorize precisely—select the most appropriate "entity" and "label".

4. Ensure valid JSON format—return structured, machine-readable output.

Your final output should be a list of JSON objects containing only the entities present in the document.
\end{promptbox}

\subsubsection{MEDEC}

\begin{promptbox}[MEDEC System Prompt]
\scriptsize
Task: Medical Error Detection in Clinical Text
Role: You are an expert medical reviewer analyzing clinical text for accuracy. Your task is to determine whether there is one medical error in the provided text.

Input Format:
The input consists of multiple sentences.
Each sentence starts with a Sentence ID, followed by the sentence itself.
Sentences are formatted one per line with a space separating the ID and the sentence text.

Types of Errors to Detect:
Diagnosis errors (incorrect or conflicting diagnoses)
Treatment errors (inappropriate or missing treatments)
Management errors (incorrect clinical decision-making)
Causation errors (incorrect understanding of disease causes or progression)

Output Format:
If one sentence contains a medical error, return only the Sentence ID of that sentence.
If no errors are found, return "-1" (without quotes).
You must not provide any explanation, only the Sentence ID or -1.

Example with an error:
input :

0 The patient was diagnosed with bacterial pneumonia and prescribed amoxicillin.  

1 The recommended treatment for viral pneumonia is antibiotics.  

2 The patient showed signs of improvement after three days.  

output:1

Example without error:

input:0 The patient was diagnosed with bacterial pneumonia and prescribed amoxicillin.  

1 The recommended treatment for bacterial pneumonia is antibiotics.

2 The patient showed signs of improvement after three days.

output:-1
\end{promptbox}

\begin{table}[h]
\small
\renewcommand{\arraystretch}{1.3}
\centering
\caption{The configuration used for CLUE+ subset datasets of the benchmark regarding few shots and the metric used. The decoding strategy on these six datasets is greedy decoding.}
\begin{tabular}{lcl}
\toprule
 \textbf{Dataset Name} & \textbf{Few-Shots} & \textbf{Metric}\\
\midrule
 ICD10CM & 3 & AVG Accuracy \\
 \hline
 \multirow{2}{*}{MedicationQA} & \multirow{2}{*}{3} & \multirow{2}{*}{\shortstack[l]{Rouge-1 F1\\\cite{lin-2004-rouge}}} \\
 &  & \\
 \hline
    \multirow{2}{*}{RRS QA} & \multirow{2}{*}{3} & \multirow{2}{*}{\shortstack[l]{Rouge-1 F1\\\cite{lin-2004-rouge}}} \\
 &  & \\
\hline
 \multirow{3}{*}{SDoH} & \multirow{3}{*}{4} & \multirow{3}{*}{\shortstack[l]{Type-Match F1\\w/ boundary overlap\\\cite{chai2018-ner-eval}}} \\
 &  & \\
 &  & \\
 \hline
\multirow{2}{*}{ACI-Bench} & \multirow{2}{*}{1} & \multirow{2}{*}{\shortstack[l]{Rouge-1 F1\\\cite{lin-2004-rouge}}} \\
 &  & \\
\hline
MEDEC & 2 & Sentence ID Accuracy \\
\bottomrule
\end{tabular}
\label{tab:cluep_detail}
\end{table}

\begin{table*}[h!]
\small
\renewcommand{\arraystretch}{1.3}
\centering
\caption{CLUE+ benchmark datasets with task types, input, and output specifications. Task types are: question-answering (QA), summarization, reasoning and information extraction (IE).}
\begin{tabular}{lllll}
\toprule
 & \textbf{Task} & \textbf{Dataset Name} & \textbf{Input} & \textbf{Output}\\
\midrule
\multirow{8}{*}{\rotatebox{90}{\scriptsize CLUE}} & {\scriptsize NLI} & MedNLI {\scriptsize \cite{shivade2017mednli}} & Premise + Hypothesis & Label \\
\cmidrule{2-5} 
     & {\scriptsize Summary} & MeQSum {\scriptsize \cite{MeQSum}} & Consumer Health Question  & Summary question \\
\cmidrule{2-5} 
    & {\scriptsize Summary} & Problem list summarization {\scriptsize \cite{gao2023bionlp}} & Progress notes & Problem list \\
\cmidrule{2-5} 
    & {\scriptsize QA} & LongHealth {\scriptsize \cite{adams2024longhealth}} & Clinical records + Question & Answer \\
\cmidrule{2-5} 
    & {\scriptsize QA} & MeDiSumQA {\scriptsize \cite{dada2025medisumqa}} & Discharge letter + Questions & Answers \\
\cmidrule{2-5} 
    & {\scriptsize Reasoning} & MeDiSumCode {\scriptsize \cite{dada2024clue}} & Discharge letter & ICD10CM codes \\
\midrule
\multirow{8}{*}{\scriptsize \rotatebox{90}{CLUE+}} & {\scriptsize QA}  & MedConceptsQA {\scriptsize ICD10CM \cite{shoham2024medconceptsqa}} & Code + definition options & Answer \\
\cmidrule{2-5} 
    & {\scriptsize QA} & MedicationQA {\scriptsize \cite{BenAbacha2019medicationqa}} & Question on medication  & Answer \\
\cmidrule{2-5} 
    & {\scriptsize QA} & MEDIQA-RRS QA {\scriptsize \cite{abacha2021mediqa}} & Findings + Questions & Answers \\
\cmidrule{2-5} 
    & {\scriptsize IE} & SDoH {\scriptsize \cite{lybarger2023n2c2}} & Clinical note & Entities \\
\cmidrule{2-5} 
    & {\scriptsize Summary} & ACI-Bench {\scriptsize \cite{yim2023aci}} & Doctor-patient dialog & Clinical note \\
\cmidrule{2-5} 
    & {\scriptsize Reasoning} & MEDEC {\scriptsize \cite{abacha2024medec}} & Clinical note & Error detection \\
\bottomrule
\end{tabular}
\label{tab:cluep_benchmark}
\end{table*}

\begin{table*}[ht!]
    \newcommand{\green}[1]{\textcolor[HTML]{008000}{#1}}
    \newcommand{\red}[1]{\textcolor{red}{#1}}
    \caption{Performances on the CLUE subset of datasets for the merged and aligned versions of MediPhi as well as other medical LLMs. PLS stands for Problem List Summary while LH refers to LongHealth.}
    \centering
    \renewcommand{\arraystretch}{1.2}
    \small
    \begin{tabular}{llcccccc}
        \toprule
        \multicolumn{2}{c}{\textbf{}} & \textbf{MedNLI} & \textbf{PLS} & \textbf{MeQSum} & \textbf{LH} & \textbf{MeDiSumQA} & \textbf{MeDiSumCode} \\ \midrule
        \textbf{Baseline} & \textit{Phi-3.5-mini-instruct} & 66.6 & 28.4 & 36.7 & 45.9 & 25.9 & 41.1 \\
        \midrule
        \multirow{6}{*}{\textbf{SLERP}} & DataMix & \green{68.5} & \green{29.0} & \green{37.7} & \red{45.7} & \green{26.6} & \green{41.4} \\ 
        \cmidrule{2-8}
         & PubMed & \green{68.3} & \green{29.2} & \green{37.6} & \red{45.7} & \green{26.3} & \red{41.0}  \\ 
         & Clinical & \green{69.2} & \green{29.4} & \green{38.1} & \red{43.5} & \green{26.7} & \red{40.5} \\ 
         & MedWiki & \green{72.8} & \green{29.2} & \green{37.6} & \red{43.6} & \red{25.1} & \green{41.7} \\ 
         & MedCode & \green{68.5} & \red{22.3} & \red{33.5} & \red{45.7} & \red{23.6} & \red{39.0} \\ 
         & Guideline & \green{70.3} & \green{29.8} & \green{37.6} & \red{41.1} & \red{25.1} & \green{41.9}\\
         \midrule
        \textbf{BreadCrumbs} & MediPhi & \green{66.9} & \green{28.8} & \green{37.9} & \red{45.7} & \green{26.1} & \green{41.7}  \\
        \midrule
        \multirow{2}{*}{\textbf{MediFlow}} & MediPhi-SFT & \green{70.6} & \red{26.9} & \green{42.8} & \red{44.2} & \green{28.8} & \red{35.0}  \\
        & MediPhi-Instruct & \green{71.0} & \red{26.0} & \green{42.8} & \red{45.0} & \green{29.1} & \red{37.2} \\
        \midrule\midrule
        \multirow{4}{*}{\small \shortstack[c]{Other\\Medical\\LLMs}} & Mistral-7B-Instruct-v0.1 & 64.8 & 25.0 & 31.1 & 30.0 & 25.5 & 13.9 \\
        & BioMistral-7B-DARE & \green{66.8} & \green{28.4} & \green{34.5} & \green{30.5} & \green{25.7} & \green{21.3} \\
        \cmidrule{2-8}
        & Meta-Llama-3-8B-Instruct & 74.1 & 31.6 & 39.5 & 58.8 & 30.3 & 27.8 \\
        & Llama3-Med42-8B & \green{77.5} & \green{32.4} & \green{42.8} & \red{57.9} & \red{29.7} & \red{25.2} \\
        \bottomrule
    \end{tabular}
    \label{tab:clue_detail_results}
\end{table*}

\begin{table*}[ht!]
    \newcommand{\green}[1]{\textcolor[HTML]{008000}{#1}}
    \newcommand{\red}[1]{\textcolor{red}{#1}}
    \caption{Performances on the new CLUE+ subset of datasets for the merged and aligned versions of MediPhi as well as other medical LLMs. ACI refers to ACI-Bench.}
    \centering
    \renewcommand{\arraystretch}{1.2}
    \small
    \begin{tabular}{llcccccc}
        \toprule
        \multicolumn{2}{c}{\textbf{}} & \textbf{RRS QA} & \textbf{MedicationQA} & \textbf{MEDEC} & \textbf{ACI} & \textbf{SDoH} & \textbf{ICD10CM} \\
        \midrule
        \textbf{Baseline} & \textit{Phi-3.5-mini-instruct} & 41.2 & 11.2 & 14.8 & 42.3 & 35.1 & 49.3 \\
        \midrule
        \multirow{6}{*}{\textbf{SLERP}} & DataMix & \green{43.3} & \red{10.8} & \green{18.8} & \green{42.7} & \green{36.2} & \green{49.5}  \\ 
        \cmidrule{2-8}
         & PubMed & \green{44.1} & \red{10.3} & \green{22.2} & \green{42.7} & \green{35.8} & \green{49.5}  \\ 
         & Clinical & \green{52.1} & \green{12.0} & \green{34.5} & \green{43.9} & \green{35.8} & \green{49.6} \\ 
         & MedWiki & \green{46.7} & \green{12.2} & \green{28.8} & \green{44.7} & \green{43.6} & \green{50.2} \\ 
         & MedCode & \green{45.6} & \green{12.0} & \green{18.1} & \red{39.0} & \red{24.8} & \green{68.7} \\ 
         & Guideline & \green{48.9} & \green{11.9} & \green{28.3} & \green{44.7} & \green{41.0} & \green{49.8}\\
         \midrule
        \textbf{BreadCrumbs} & MediPhi & \green{44.5} & \green{11.3} & \green{29.1} & \green{44.3} & \green{39.7} & \green{55.5}  \\
        \midrule
        \multirow{2}{*}{\textbf{MediFlow}} & MediPhi-SFT & \green{60.8} & \green{18.8} & \green{35.0} & \green{43.4} & \green{54.5} & \green{54.9}  \\
        & MediPhi-Instruct & \green{61.6} & \green{19.3} & \green{34.4} & \green{43.5} & \green{56.7} & \green{54.9} \\
        \midrule\midrule
        \multirow{4}{*}{\small \shortstack[c]{Other\\Medical\\LLMs}} & Mistral-7B-Instruct-v0.1 & 50.4 & 22.7 & 21.5 & 50.4 & 40.2 & 27.6 \\
        & BioMistral-7B-DARE & \red{49.6} & \red{22.3} & \green{23.1} & \red{43.3} & \green{45.9} & \red{25.1} \\
        \cmidrule{2-8}
        & Meta-Llama-3-8B-Instruct & 55.8 & 26.1 & 46.5 & 50.2 & 63.1 & 25.7 \\
        & Llama3-Med42-8B & \red{54.1} & \red{25.7} & \red{35.4} & \green{56.5} & \red{53.9} & \green{53.4} \\
        \bottomrule
    \end{tabular}
    \label{tab:cluep_detail_results}
\end{table*}

\begin{table*}[ht!]
    \newcommand{\green}[1]{\textcolor[HTML]{008000}{#1}}
    \newcommand{\red}[1]{\textcolor{red}{#1}}
    \caption{Performances of MediPhi models on multiple-choice question-answering medical benchmarks.}
    \centering
    \renewcommand{\arraystretch}{1.2}
    \small
    \begin{tabular}{lccccc}
        \toprule
         & \textbf{MedQA} & \textbf{MedMCQA} & \textbf{PubMedQA} & \textbf{MMLU-med} & \textbf{AVG} \\
        \midrule
        \textit{Phi-3.5-mini-instruct} & 0.486 & 0.554 & 0.768 & 0.715 & 0.631 \\
        \midrule
        PubMed & \red{0.467} & \red{0.549} & \green{0.774} & \green{0.724} & \red{0.629} \\
         Clinical & \green{0.500} & \red{0.551} & \green{0.772} & \green{0.727} & \green{0.638} \\
         Medical & \green{0.519} & \red{0.535} & \red{0.740} & \red{0.701} & \red{0.624} \\
         MedWiki & \red{0.478} & \red{0.548} & 0.768 & \green{0.719} & \red{0.628} \\
         MedCode & \green{0.531} & \red{0.532} & \red{0.758} & \red{0.700} & \red{0.630} \\
         Guideline & \red{0.459} & \green{0.556} & \green{0.770} & \green{0.724} & \red{0.627} \\
         \midrule
        MediPhi & \green{0.491} & \green{0.559} & \red{0.766} & \green{0.720} & \green{0.634} \\
        \midrule
        MediPhi-SFT & \green{0.536} & \red{0.552} & \red{0.766} & \green{0.716} & \green{0.642} \\
        MediPhi-Instruct & \green{0.548} & \green{0.555} & \red{0.764} & \red{0.714} & \green{0.645} \\
        \bottomrule
    \end{tabular}
    \label{tab:trad_results}
\end{table*}

\end{document}